\documentclass[acmlarge,screen]{acmart}
\usepackage{multirow}
\usepackage{tabularx}
\usepackage{array}
\usepackage{longtable}
\usepackage{enumitem}
\usepackage{wasysym}
\usepackage{bm}
\usepackage{algorithm}
\usepackage{algpseudocode}

\AtBeginDocument{}

\setcopyright{none}
\copyrightyear{2026}
\acmYear{2026}
\acmDOI{XXXXXXX.XXXXXXX}
\acmJournal{CSUR}
\acmVolume{1}
\acmNumber{1}
\acmArticle{1}
\acmMonth{1}

\begin{document}

\title[Trustworthy Knowledge Distillation in Federated Learning]{Trustworthy Knowledge Distillation in Federated Learning: A Systematic Review of Proxy Strategies}

\author{Agostinho Freire}
\orcid{0000-0002-6059-9014}
\email{aafj@ecomp.poli.br}
\affiliation{%
  \institution{Escola Polit\'ecnica de Pernambuco, Universidade de Pernambuco (UPE)}
  \city{Recife}
  \state{Pernambuco}
  \country{Brazil}}
 
\author{Jo\~ao V. R. Andrade}
\orcid{0000-0003-3554-0157}
\email{jvlc@ecomp.poli.br}
\affiliation{%
  \institution{Escola Polit\'ecnica de Pernambuco, Universidade de Pernambuco (UPE)}
  \city{Recife}
  \state{Pernambuco}
  \country{Brazil}}
\affiliation{%
  \institution{L3S Research Center, Leibniz Universit\"at Hannover}
  \city{Hannover}
  \country{Germany}}
 
\author{Leandro H. de S. Silva}
\orcid{0000-0001-6221-2250}
\email{lhss@ecomp.poli.br}
\affiliation{%
  \institution{Escola Polit\'ecnica de Pernambuco, Universidade de Pernambuco (UPE)}
  \city{Recife}
  \state{Pernambuco}
  \country{Brazil}}
\affiliation{%
  \institution{Campus Cajazeiras, Instituto Federal do Sert\~ao da Para\'iba (IFSPB)}
  \city{Cajazeiras}
  \state{Para\'iba}
  \country{Brazil}}
 
\author{Juan Lira}
\orcid{0009-0005-6990-9532}
\email{jcml@poli.br}
\affiliation{%
  \institution{Escola Polit\'ecnica de Pernambuco, Universidade de Pernambuco (UPE)}
  \city{Recife}
  \state{Pernambuco}
  \country{Brazil}}
 
\author{Marco Fisichella}
\orcid{0000-0002-6894-1101}
\email{mfisichella@l3s.de}
\affiliation{%
  \institution{L3S Research Center, Leibniz Universit\"at Hannover}
  \city{Hannover}
  \country{Germany}}
 
\author{Bruno J. T. Fernandes}
\orcid{0000-0002-6001-3925}
\email{bruno.fernandes@l3s.uni-hannover.de}
\affiliation{%
  \institution{Escola Polit\'ecnica de Pernambuco, Universidade de Pernambuco (UPE)}
  \city{Recife}
  \state{Pernambuco}
  \country{Brazil}}
\affiliation{%
  \institution{L3S Research Center, Leibniz Universit\"at Hannover}
  \city{Hannover}
  \country{Germany}}

\renewcommand{\shortauthors}{Freire et al.}

\begin{abstract}
Knowledge distillation is a communication-efficient alternative to
parameter averaging in federated learning, but it requires a shared
non-private proxy dataset.  We systematically review how such proxies
are constructed, spanning real public references and synthetic
alternatives.  Following a PRISMA~2020-aligned protocol, we analyze 51
primary studies (2020--2025) along three axes: proxy mechanism,
distillation flow, and ownership scope.  We find that 77\% of studies use
generated, synthetic, or hybrid proxies; that reporting on
communication, privacy, and robustness is uneven; and that none
releases its synthetic reference data.  We derive five design
recommendations and a seven-direction research agenda for trustworthy
federated distillation.
\end{abstract}



\maketitle

\section{Introduction}
\label{sec:introduction}

The volume and variety of data generated at the edges of modern
networks---by mobile devices, hospital information systems,
industrial sensors, and financial platforms---have grown far beyond
what centralized training pipelines can safely absorb.  Legislation
such as the GDPR, HIPAA, Brazil's LGPD, and China's PIPL restricts
cross-border data movement, mandates data minimization, and grants
individuals rights over their personal information; in healthcare
alone, more than 133~million records were exposed in breaches in
2023, underscoring the practical urgency of privacy-preserving
alternatives to centralized learning~\cite{kairouz2021advances,
singh2023safeguard}.

Federated learning (FL) was introduced precisely to address this
tension~\cite{mcmahan2017fedavg}: training data stays on each
participant's device, and collaboration proceeds through iterative
exchange of model updates, as in the canonical \textsc{FedAvg}
protocol.  Despite its appeal, parameter-averaging FL faces three
persistent, interrelated challenges: the \emph{communication cost}
of transmitting full parameter vectors at every round over
constrained links~\cite{kairouz2021advances}; \emph{statistical
heterogeneity}, since real client data is almost never IID---label,
feature, and quantity skew induce \emph{client drift} and degrade
the averaged global model~\cite{hsu2019measuring, li2020fedprox,
zhao2018federated}; and \emph{architectural homogeneity}, because
element-wise averaging requires every client to run the same model,
which resource-diverse participants cannot~\cite{kairouz2021advances,
li2019fedmd}.

Knowledge distillation~\cite{hinton2015distilling} offers a
principled way to sidestep all three bottlenecks simultaneously.
Applied in a federated context, distillation replaces weight
exchange with the exchange of predictions---logits, soft labels, or
feature representations---computed over a common reference
set~\cite{jeong2018communication, li2019fedmd, lin2020ensemble}.
This yields communication that scales with classes and reference
samples rather than parameters, natural support for heterogeneous
client architectures, and a measure of privacy amplification, since
soft predictions are a lossy one-way function of the private
data~\cite{abadi2016deep, dwork2006differential}.  Federated
distillation was first explored by Jeong et
al.~\cite{jeong2018communication}; Lin et al.~\cite{lin2020ensemble}
formalized ensemble distillation for robust model fusion; and
FedMD~\cite{li2019fedmd} enabled heterogeneous architectures by
distilling over a shared public dataset.  These foundational works
demonstrated the paradigm's potential---and revealed its Achilles'
heel: the dependence on a suitable \emph{proxy dataset}.

Nearly every distillation-based federated method assumes a shared,
non-private reference set---the \emph{proxy}---on which participants
compute and exchange predictions.  ``Non-private'' here is the broad
property that the proxy may be shared among all participants without
violating the privacy constraints that motivated the federation: it
spans both real public references (a pre-existing public dataset
reused unaltered) and synthetic data generated for the federation.
A public dataset is therefore one particular kind of non-private
proxy, not a synonym for it---a distinction central to the scope of
this review.  In practice, the idealized assumptions behind the
proxy are fragile on three fronts.  \emph{Domain mismatch}: when the
proxy is a general-purpose public dataset, the distribution gap
between proxy and private data degrades the distilled knowledge.
\emph{Legal and ethical risk}: in regulated domains, even curating a
small shared dataset can be impractical, and a proxy legal in one
jurisdiction may violate data residency in another.
\emph{Availability}: for many specialized tasks---rare-disease
diagnosis, industrial fault detection, proprietary network
traffic---no suitable public reference exists.  These limitations
are not theoretical: in our corpus of 51 included studies, 71\,\% of
the methods that address the proxy problem do so by generating
synthetic data or using data-free inversion precisely because a
suitable public proxy is unavailable or inadequate.

A rapidly growing body of work therefore produces \emph{synthetic
proxies}: reference inputs generated by non-private
mechanisms---GAN-based generation~\cite{goodfellow2014generative} (the
modal mechanism, 25\,\% of included studies), diffusion
models~\cite{ho2020denoising}, data-free deep
inversion~\cite{yin2020dreaming, zhu2021datafree}, VAE and
autoencoder schemes, and curated or hybrid combinations.  Despite
this proliferation---73\,\% of the included studies were published
in 2024 or 2025---the literature remains fragmented in terminology
(``transfer set,'' ``reference data,'' ``pseudo-data,''
``distillation buffer''), in non-IID simulation strategies, in the
metrics reported, and in reproducibility: only 22\,\% of included
studies release code, and only 4\,\% make their synthetic
references available even partially.

Existing surveys provide broad FL overviews~\cite{kairouz2021advances,
li2020federated_survey} or focus on sub-problems such as non-IID
data~\cite{zhao2018federated} and knowledge distillation in
general~\cite{gou2021knowledge}, but none systematically maps and
critically evaluates the intersection of synthetic proxy generation
and distillation-based aggregation in the federated setting.  The
absence of a systematic review leaves practitioners without
consolidated guidance on which proxy mechanism to adopt, which
quality criteria to monitor, and which experimental practices to
follow---and leaves recurring gaps (bytes-to-convergence omission,
scarce robustness evaluation, opaque proxy-quality assessment)
undiagnosed.  This systematic literature review fills this gap.

\subsection{Scope and methodology overview}
\label{sec:intro_scope}

We follow Kitchenham and Charters~\cite{kitchenham2007guidelines}
and the PRISMA~2020 reporting standard~\cite{page2021prisma}.  A
registered protocol specifies six research questions covering:
(RQ1)~how synthetic proxies are used within the client--server
flow; (RQ2)~which generation mechanisms and families appear;
(RQ3)~what effects are reported on communication, privacy, and
performance under non-IID settings; (RQ4)~which proxy-quality
criteria are reported and how they relate to outcomes; (RQ5)~which
threat models, robustness analyses, and defenses are evaluated; and
(RQ6)~which methodological gaps persist.  The scope is the
intersection of three conditions: the study must (i)~target
federated learning, (ii)~replace or augment parameter exchange with
prediction- or representation-level transfer, and (iii)~use a
synthetic, generated, or curated non-private reference as the proxy.
We searched five databases (IEEE Xplore, ACM Digital Library,
Scopus, ScienceDirect, Springer Link) from 2019 to 2025 with a
field-aware query; of 501 retrieved records, staged screening
retained 51 primary studies for full extraction, with a mean quality
score of 6.7/8.0 on an eight-item checklist.

\subsection{Contributions}
\label{sec:intro_contributions}

The contributions of this review are fourfold:

\begin{enumerate}[leftmargin=*]
  \item \textbf{Protocol-driven, transparent review.}  A fully
    documented, reproducible systematic review of distillation-based
    FL across the full spectrum of non-private proxies, covering 51
    primary studies (2019--2025) in 33~journals and 17~conferences,
    with all search strings, screening decisions, quality scores,
    and extraction fields recorded and available.
  \item \textbf{Multi-axis taxonomy.}  A taxonomy along three axes:
    (a)~proxy generation mechanism; (b)~distillation and aggregation
    design (flow direction, prediction object, ownership scope); and
    (c)~heterogeneity treatment (non-IID type, client count, FL
    setting, domain).
  \item \textbf{Cross-dimensional evidence synthesis.}  A
    cross-walk of proxy-quality indicators against downstream
    outcomes, a consolidation of reporting practices for
    communication, privacy, and robustness, and a quantification of
    the transparency gaps (22\,\% code release, 4\,\% synthetic
    references, 67\,\% with at most limited privacy discussion).
  \item \textbf{Actionable research agenda.}  Five persistent gaps
    and concrete recommendations: standardized non-IID benchmarks;
    mandatory bytes-to-convergence reporting; robustness evaluation
    under noisy, malicious, and out-of-distribution proxy
    conditions; transparent release of synthetic proxies; and formal
    proxy-quality metrics linked to downstream performance.
\end{enumerate}

Taken together, these contributions frame the use of synthetic and
public proxies as a question of \emph{trustworthiness}: a federated
distillation system is only as trustworthy as the proxy it relies on
is privacy-preserving, robust to unreliable or adversarial
participants, and reproducible by independent parties.  Throughout
the review we therefore read the evidence not only through
predictive performance, but through the three trustworthiness
dimensions of privacy (RQ3), robustness (RQ5), and reproducibility
(RQ6), which jointly determine whether a method can be responsibly
deployed in the regulated settings that motivate federated learning
in the first place.

\subsection{Organization}

Section~\ref{sec:related} positions the review against existing
surveys; Section~\ref{sec:background} provides the conceptual
background; Section~\ref{sec:methodology} details the protocol;
Section~\ref{sec:results} presents the synthesis around the six RQs;
Sections~\ref{sec:discussion} and~\ref{sec:future} discuss
implications and future directions; Section~\ref{sec:conclusion}
concludes.

\section{Related Surveys and Positioning}
\label{sec:related}

Several surveys touch on the topics of this review, yet none
addresses the specific intersection of federated learning, knowledge
distillation, and synthetic proxies under a systematic,
protocol-driven methodology.  We organize the closest literature
into four groups; Table~\ref{tab:related_surveys} consolidates the
comparison.

\textbf{General FL surveys.}  The foundational survey of Kairouz et
al.~\cite{kairouz2021advances} remains the canonical FL reference
but treats distillation only briefly and synthetic proxies not at
all.  More recent surveys extend the perspective in different
directions---design and functional models of
FL~\cite{ayeelyan2024fl_survey}, a systematic review of 238 studies
from the clients' perspective~\cite{yu2024client_perspective}, data
security and privacy in edge-IoT deployments~\cite{xu2024fl_iot_survey},
and a PRISMA-based SLR on federated NLP~\cite{manoj2024fl_nlp_slr}.
These works converge on non-IID data, communication cost, and
privacy as the core FL challenges, but none analyzes how synthetic
proxies address those challenges through distillation.

\textbf{Knowledge-distillation surveys.}  Gou et
al.~\cite{gou2021knowledge} provide the most cited KD survey,
organized by knowledge type, distillation scheme, and
teacher--student architecture; Moslemi et
al.~\cite{moslemi2024kd_survey} offer a recent overview with a
dedicated treatment of data-free KD, adversarial distillation, and
cross-modal transfer, and Liu et al.~\cite{liu2024dfkd_cvpr} analyze
the quality and diversity of DFKD-generated samples.  These works
treat KD as a centralized compression and transfer technique: their
DFKD discussion centers on the single-teacher, single-student
paradigm, without multi-client aggregation, non-IID data, or
communication budgets.

\textbf{KD-in-FL surveys.}  The closest works focus explicitly on
the intersection.  Wu et al.~\cite{wu2024fel_survey}, in a preprint
first released in 2023 and most recently revised in 2024, survey KD
in federated edge learning, organized by FL challenge; Qin et
al.~\cite{qin2024fedkd_survey} discuss KD as an alternative to
parameter-based FL across heterogeneous-model, personalized, and
class-incremental settings; and an IJCAI survey-track practical
guide~\cite{ijcai2024fedkd_practical} classifies KD-in-FL by target
problem with pseudo-code.  Our review differs in three respects:
\emph{scope}---they treat synthetic or data-free proxies as one
strategy among many, whereas we make the proxy mechanism the primary
taxonomic axis and analyze its interaction with distillation flow
and heterogeneity treatment; \emph{methodology}---they are narrative,
whereas we follow Kitchenham/PRISMA with a registered protocol,
predefined criteria, and a uniform quality checklist, yielding a
reproducible corpus; and \emph{diagnostic focus}---we explicitly
measure reporting and reproducibility gaps (code release, synthetic
references, bytes-to-convergence, robustness evaluation), which, to
our knowledge, no previous review has quantified in this subfield.

\begin{table*}[!htbp]
  \centering
  \caption{Comparison of our review to the closest related surveys.
    \textbf{FL}: federated learning coverage; \textbf{KD}: knowledge
    distillation coverage; \textbf{Synth.\ proxy}: explicit focus on
    synthetic/generated proxies as a design axis;
    \textbf{Systematic}: protocol-driven (PRISMA~2020 or Kitchenham)
    methodology; \textbf{QA}: formal quality assessment of the
    included studies; \textbf{Reprod.}: quantitative diagnosis of
    the reproducibility gap (code and synthetic-reference release).}
  \label{tab:related_surveys}
  \scriptsize
  \setlength{\tabcolsep}{4pt}
  \renewcommand{\arraystretch}{1.1}
  \begin{tabularx}{\textwidth}{@{}>{\raggedright\arraybackslash}Xccccccc@{}}
    \toprule
    \textbf{Survey} & \textbf{Year} & \textbf{FL} & \textbf{KD} &
    \textbf{Synth.\ proxy} & \textbf{Systematic} & \textbf{QA} &
    \textbf{Reprod.}                                                \\
    \midrule
    Kairouz et al.~\cite{kairouz2021advances}              & 2021 & \CIRCLE & \LEFTcircle & $-$         & $-$         & $-$ & $-$ \\
    Gou et al.~\cite{gou2021knowledge}                     & 2021 & $-$      & \CIRCLE     & \LEFTcircle & $-$         & $-$ & $-$ \\
    Yu et al.~\cite{yu2024client_perspective}              & 2023 & \CIRCLE & $-$         & $-$         & \CIRCLE     & \LEFTcircle & $-$ \\
    Wu et al.~\cite{wu2024fel_survey}                      & 2024 & \CIRCLE & \CIRCLE     & \LEFTcircle & $-$         & $-$ & $-$ \\
    Qin et al.~\cite{qin2024fedkd_survey}                  & 2024 & \CIRCLE & \CIRCLE     & \LEFTcircle & $-$         & $-$ & $-$ \\
    Moslemi et al.~\cite{moslemi2024kd_survey}             & 2024 & $-$      & \CIRCLE     & \LEFTcircle & $-$         & $-$ & $-$ \\
    IJCAI practical guide~\cite{ijcai2024fedkd_practical}  & 2024 & \CIRCLE & \CIRCLE     & \LEFTcircle & $-$         & $-$ & $-$ \\
    Ayeelyan et al.~\cite{ayeelyan2024fl_survey}           & 2024 & \CIRCLE & \LEFTcircle & $-$         & $-$         & $-$ & $-$ \\
    \midrule
    \textbf{This review} (2026) & \textbf{2026} & \CIRCLE & \CIRCLE & \CIRCLE & \CIRCLE & \CIRCLE & \CIRCLE \\
    \bottomrule
  \end{tabularx}
\end{table*}

In short, this review is unique along four simultaneous dimensions:
a narrow focus on the three-way intersection of FL, distillation,
and synthetic proxies; a registered, PRISMA-based protocol; a formal
quality assessment applied uniformly to the 51 included studies; and
a quantitative diagnosis of the reproducibility and reporting gaps,
which grounds the research agenda of Section~\ref{sec:future}.
These four dimensions justify a new systematic review and
differentiate it from the narrative surveys above.

\section{Background}
\label{sec:background}

This section formalizes the federated learning setting
(Section~\ref{sec:bg_fl}), knowledge distillation and its
specialization to distributed settings (Section~\ref{sec:bg_kd}),
and the families of synthetic and public proxies that structure the
taxonomy of Section~\ref{sec:results}
(Section~\ref{sec:bg_proxies}).  Readers familiar with FL and KD may
proceed directly to Section~\ref{sec:bg_proxies}.

\subsection{Federated learning}
\label{sec:bg_fl}

In federated learning, $K$ clients collaborate to train a shared
model without centralizing their private data.  Let client $k$ hold
a local dataset $\mathcal{D}_k$ of size $n_k$, with
$n = \sum_{k=1}^{K} n_k$.  The global objective is the weighted
average of client-specific empirical risks:

\begin{equation}
  \label{eq:fl_objective}
  \min_{\bm{\theta} \in \mathbb{R}^d} \;
    F(\bm{\theta})
  \;=\;
    \sum_{k=1}^{K} \frac{n_k}{n} \, F_k(\bm{\theta}),
  \qquad
    F_k(\bm{\theta})
  \;=\;
    \frac{1}{n_k}
    \sum_{(x, y) \in \mathcal{D}_k}
      \ell\!\left(\bm{\theta}; x, y\right),
\end{equation}

where $\bm{\theta} \in \mathbb{R}^d$ are the model parameters and
$\ell$ a per-sample loss~\cite{mcmahan2017fedavg,
kairouz2021advances}.  The canonical algorithm,
\textsc{FedAvg}~\cite{mcmahan2017fedavg}, solves
\eqref{eq:fl_objective} in communication rounds: the server
broadcasts the current parameters to a selected subset
$\mathcal{S}_t$ of clients, each performs $E$ local epochs of SGD,
and the server aggregates by weighted averaging:

\begin{equation}
  \label{eq:fedavg_aggregation}
  \bm{\theta}^{(t+1)} \;=\;
  \sum_{k \in \mathcal{S}_t}
    \frac{n_k}{\sum_{j \in \mathcal{S}_t} n_j} \,
    \bm{\theta}_k^{(t, E)}.
\end{equation}

This recipe transmits each parameter vector once per round,
amortized over $E$ local epochs, and preserves privacy in the weak
sense that the server never observes raw data.  Its elegance,
however, masks three challenges that motivate the distillation-based
approaches reviewed here.  \emph{Statistical heterogeneity and
client drift}: under non-IID conditions (label, feature, or quantity
skew, or temporal drift~\cite{ayeelyan2024fl_survey,
hsu2019measuring, zhao2018federated}), each local objective has its
own minimizer, local SGD drives clients toward client-specific
optima, and the aggregation step becomes biased---the global model
oscillates, converges slowly, or plateaus well below the centralized
baseline~\cite{li2020fedprox, yu2024client_perspective}.
\emph{Communication cost}: transmitting the full parameter vector
each round is a first-order bottleneck in cross-device deployments
with constrained links~\cite{kairouz2021advances, xu2024fl_iot_survey};
compression mitigates but complicates optimization, whereas
transmitting class logits per proxy sample is orders of magnitude
smaller.  \emph{Architectural homogeneity}: element-wise averaging
presumes identical architectures, which resource-diverse
participants cannot satisfy; prediction-level transfer is the
natural alternative~\cite{li2019fedmd, ijcai2024fedkd_practical}.

Two main FL settings are conventionally
distinguished~\cite{kairouz2021advances}: \textbf{cross-device} FL,
with very large populations ($10^3$--$10^{10}$) of
resource-constrained devices holding small datasets under
intermittent connectivity, and \textbf{cross-silo} FL, with few
institutional participants (2 to $\approx$100) holding larger
datasets under stringent regulatory constraints; hybrid and
hierarchical variants combine the two via edge
servers~\cite{wu2024fel_survey}.  As Section~\ref{sec:results}
shows, most included studies target cross-device FL with mixed
heterogeneity, while cross-silo settings dominate healthcare and
industry.

\subsection{Knowledge distillation in distributed settings}
\label{sec:bg_kd}

Knowledge distillation (KD) was proposed for model
compression~\cite{hinton2015distilling, gou2021knowledge}: a teacher
$f_{\bm{\theta}_T}$ transfers knowledge to a student
$f_{\bm{\theta}_S}$ by having the student match the teacher's
softened output distribution.  Given teacher logits
$\bm{z}_T(x)$, the softened distribution at temperature
$\tau \ge 1$ is

\begin{equation}
  \label{eq:softmax_temperature}
  p_T^{(\tau)}(x)_c
  \;=\;
  \frac{\exp\!\left(z_T(x)_c \,/\, \tau\right)}
       {\sum_{c'=1}^{C} \exp\!\left(z_T(x)_{c'} \,/\, \tau\right)},
  \qquad c = 1, \dots, C.
\end{equation}

Temperatures $\tau > 1$ amplify non-dominant classes, exposing the
\emph{dark knowledge} in inter-class similarities.  The student
minimizes a convex combination of a distillation loss and standard
cross-entropy:

\begin{equation}
  \label{eq:kd_loss}
  \mathcal{L}_{\mathrm{KD}}(x, y)
  \;=\;
  (1 - \alpha) \,
    \ell_{\mathrm{CE}}\!\left(p_S^{(1)}(x), y\right)
  \;+\;
  \alpha \, \tau^2 \,
    \mathrm{KL}\!\left(
      p_T^{(\tau)}(x) \,\big\|\, p_S^{(\tau)}(x)
    \right),
\end{equation}

with balancing hyperparameter $\alpha \in [0,1]$; recent surveys
consolidate response-, feature-, and relation-based variants and
extensions to self-, online, and cross-modal
distillation~\cite{hinton2015distilling, moslemi2024kd_survey}.

In the federated setting, the teacher--student dichotomy
generalizes: multiple clients act as teachers of a global student
(ensemble distillation~\cite{lin2020ensemble}), the server acts as
teacher toward clients, or transfer is bidirectional within a round.
The key primitive is the exchange of predictions computed on a
common reference---the proxy.  Let
$\mathcal{D}_{\mathrm{prx}} = \{x_1, \ldots, x_M\}$ be a proxy of
$M$ unlabeled inputs accessible to every participant.  In
client-to-server distillation, client $k$ transmits its softened
predictions on every proxy sample, and the server aggregates them
into a consensus target

\begin{equation}
  \label{eq:consensus_target}
  \bar{p}^{(\tau)}(x_i)
  \;=\;
  \sum_{k \in \mathcal{S}_t}
    w_k \,
    p_k^{(\tau)}(x_i),
  \qquad
  \sum_{k \in \mathcal{S}_t} w_k = 1,
\end{equation}

used to update the global model by minimizing
\eqref{eq:kd_loss} on $\mathcal{D}_{\mathrm{prx}}$; the weights
$w_k$ may be uniform, proportional to dataset sizes, or
adaptive~\cite{qin2024fedkd_survey}.  Two design axes characterize a
method within this framework: the \emph{distillation object} (logits,
softened probabilities, intermediate features, or prototype
embeddings, trading communication cost against knowledge richness)
and the \emph{distillation flow} (client-to-server, server-to-client,
bidirectional, or local-only).  Relative to parameter averaging,
distillation bounds communication by $\mathcal{O}(MC)$ rather than
$\mathcal{O}(d)$, supports heterogeneous architectures natively, and
provides weak privacy amplification that can be strengthened with
differential privacy~\cite{abadi2016deep, dwork2006differential}---at
the cost of a new central dependency: the proxy set on which all
predictions are computed.

\subsection{Synthetic proxies: definitions and families}
\label{sec:bg_proxies}

A \emph{proxy set} is any shared collection of reference inputs used
to compute, aggregate, or filter predictions during federated
distillation.  Its defining characteristic is being
\emph{non-private}: every participant can access it without
violating the privacy constraints that motivated the federation.
Proxies differ along three practical dimensions:

\begin{description}[style=nextline,font=\normalfont\bfseries,leftmargin=1em,itemsep=2pt]
  \item[Source] --- real curated samples, artificially generated
    samples, or a combination of both;
  \item[Alignment] --- the extent to which the proxy distribution
    matches the clients' private distribution, from in-domain to
    off-domain;
  \item[Ownership scope] --- globally shared, server-maintained,
    generated locally at each client, or shared within groups.
\end{description}

The reviewed literature organizes proxy mechanisms into six
families, which are not mutually exclusive---several included
studies combine two or more.  A further technique deserves mention
as a distinct proxy-construction family, though only marginally
represented in the corpus: \emph{dataset distillation} (dataset
condensation) compresses a training set into a few highly
informative synthetic samples by optimizing the samples themselves
rather than a generator~\cite{wang2018dataset, zhao2021dataset}; one
included study communicates synthetic \emph{distillates} of this
kind~\cite{zhang2024one}, and its systematic exploration in the
federated setting remains largely open.

\textbf{Curated and public proxies.}  The simplest family assumes a
pre-collected public dataset shareable without privacy
concerns---typically a small unlabeled slice of a standard
benchmark---as in the seminal FedMD~\cite{li2019fedmd} and
FedDF~\cite{lin2020ensemble}.  Inexpensive and deterministic, but
contingent on a suitable, legally shareable public dataset existing,
a condition frequently violated in healthcare, industrial, and
cross-border deployments.

\textbf{GAN-based generation.}  A generator--discriminator pair
trained adversarially~\cite{goodfellow2014generative}, in conditional
(class-balanced synthesis under label skew), auxiliary-classifier,
and Wasserstein-with-gradient-penalty variants; the GAN may be
trained on the server, per client, or federatedly.  Versatile and
tunable for balance and diversity, but subject to training
instability and to memorizing private information without
safeguards.

\textbf{Diffusion-based generation.}  Denoising diffusion
probabilistic models~\cite{ho2020denoising} learn to reverse a gradual
noising process and typically surpass GANs in fidelity and training
stability; the literature explores server-side diffusion and
federated diffusion, often truncating the reverse process to trade
fidelity against generation time~\cite{moslemi2024kd_survey}.

\textbf{Deep inversion.}  Generator-free synthesis that optimizes
random noise to match the statistics stored in a trained
model---most notably batch-normalization running
statistics---under a composite objective of class alignment,
BN-statistics matching, and image priors~\cite{yin2020dreaming,
zhu2021datafree}:

\begin{equation}
  \label{eq:deep_inversion}
  \hat{x}^{\star}
  \;=\;
  \arg\min_{\hat{x}} \;
    \underbrace{\ell_{\mathrm{CE}}\!\left(f_{\bm{\theta}}(\hat{x}), y\right)}_{\text{class alignment}}
  \;+\;
  \underbrace{\lambda_{\mathrm{BN}} \, \mathcal{R}_{\mathrm{BN}}(\hat{x})}_{\text{BN-statistics matching}}
  \;+\;
  \underbrace{\lambda_{\mathrm{TV}} \, \mathcal{R}_{\mathrm{TV}}(\hat{x})}_{\text{image prior}},
\end{equation}

Attractive in federated settings because it requires no real data
and no additional generator, but costly per synthesized sample and
prone to low diversity without auxiliary objectives.

\textbf{VAE and autoencoder-based schemes.}  Variational
autoencoders~\cite{kingma2014vae} and conditional
variants~\cite{sohn2015cvae} offer more stable training than GANs
and cheaper sampling than diffusion; in the reviewed literature they
appear for low-dimensional and tabular/time-series synthesis---as in
PeFAD for multivariate anomaly detection~\cite{xu2024pefad} and
Fed2KD for pandemic risk assessment~\cite{sun2022fed2kd}---and in
combination with core-set selection and latent perturbation, as in
the one-shot distillate framework of Zhang et
al.~\cite{zhang2024one}.

\textbf{Hybrid and dynamic proxies.}  Combinations of the above:
public seeds augmented with synthetic samples for under-represented
classes, GAN candidates filtered by confidence against the current
global model, or core-set selection followed by Fourier- or
latent-space perturbation.  Hybrids trade conceptual simplicity for
flexibility and often achieve the best empirical trade-offs at the
cost of additional hyperparameters.

\subsection{Taxonomy axes}
\label{sec:bg_taxonomy}

The thematic synthesis of Section~\ref{sec:results} is organized
along three axes: \textbf{(A1)~proxy mechanism}
(Section~\ref{sec:bg_proxies}); \textbf{(A2)~distillation and
aggregation design}---object, flow direction, and ownership scope
(Section~\ref{sec:bg_kd}); and \textbf{(A3)~heterogeneity
handling}---FL setting, non-IID type, client count, and domain
(Section~\ref{sec:bg_fl}).  The axes are orthogonal in principle but
strongly correlated in practice, as the cross-tabulations of
Section~\ref{sec:results} make explicit.

\section{Methodology}
\label{sec:methodology}

This review follows a protocol-driven process aligned with
Kitchenham and Charters~\cite{kitchenham2007guidelines} and
PRISMA~2020~\cite{page2021prisma}.  The protocol was registered
before any screening activity and specifies the review's objectives,
research questions, data sources and time window, search string,
eligibility criteria, selection procedure, quality checklist, and
extraction and synthesis plan.  Every decision is traceable to
explicit evidence in the source documents (page, figure, or table),
and the full extraction workbook---including the consolidated
per-study tables, all categorical fields, raw textual extractions,
and per-item quality scores---is publicly archived on Zenodo with a
persistent DOI~\cite{freire2026workbook}.

\subsection{Objectives}
\label{sec:method_objectives}

The general objective is to identify, classify, and critically
evaluate knowledge-distillation methods in federated learning that
rely on synthetic or otherwise non-private proxy data, assessing
their effects on communication efficiency, privacy preservation,
and predictive performance under heterogeneous client data.  Four
specific objectives decompose it: \textbf{(O1)}~map the generative
and curation mechanisms used to produce proxy data, with their
sampling and selection policies; \textbf{(O2)}~relate reported
proxy-quality indicators (diversity, realism, coverage, alignment)
to performance, convergence, and stability; \textbf{(O3)}~identify
the strategies for aggregating and filtering soft targets from
heterogeneous clients; and \textbf{(O4)}~surface methodological
gaps, unstandardized protocols, and transparency deficits.

\subsection{Research questions}
\label{sec:method_rqs}

Six research questions operationalize these objectives, each
answered through a distinct extraction instrument and feeding a
specific part of Section~\ref{sec:results}:

\begin{description}[style=nextline,font=\normalfont\bfseries,leftmargin=1em,itemsep=2pt]
  \item[RQ1 --- Proxy use.]  How are synthetic or generated data
    used as a proxy for distillation in FL, and what is the proxy's
    role within the client--server flow?
  \item[RQ2 --- Proxy generation.]  Which generation or selection
    techniques (GANs, diffusion, deep inversion, curated subsets,
    hybrids) are employed, and in which domains?
  \item[RQ3 --- Effects.]  What effects are reported on
    communication cost, privacy guarantees or assumptions, and
    predictive performance under non-IID data?
  \item[RQ4 --- Proxy quality.]  Which proxy-quality criteria and
    metrics are reported, and how do they relate to downstream
    performance and stability?
  \item[RQ5 --- Threats and defenses.]  Which threat models (noisy
    clients, malicious participants, out-of-distribution proxies,
    poisoned logits) and defense or filtering strategies are
    evaluated?
  \item[RQ6 --- Methodological gaps.]  Which gaps remain in non-IID
    protocols, benchmark adequacy, transparency, and
    reproducibility?
\end{description}

\subsection{Data sources, time window, and search string}
\label{sec:method_sources}

Searches covered five databases that jointly span the relevant
computer-science and engineering literature: IEEE Xplore, ACM
Digital Library, Scopus, ScienceDirect, and Springer~Link.  Preprint
servers and other grey literature (e.g., arXiv) were deliberately
excluded: restricting the corpus to venues with formal peer review
acts as a baseline quality filter, consistent with the
trustworthiness focus of this review; the cost of this choice---the
omission of very recent unrefereed work in a fast-moving field---is
acknowledged as a threat to external validity in
Section~\ref{sec:method_threats}.  The time window spans
\textbf{2019--2025}; the lower bound includes the earliest
non-trivial combinations of FL and KD---most notably Jeong et
al.~\cite{jeong2018communication}---while the bulk of the
literature has emerged since 2020.  The query returned 501 records,
with ScienceDirect (151), Scopus (122), and ACM~DL (119)
contributing the largest shares, and Springer~Link (69) and
IEEE~Xplore (40) the smallest (Supplementary Fig.~S1).

A field-aware search string requires \texttt{"federated learning"}
in the \emph{title} and, in the abstract, both a distillation-related
and a synthetic-proxy-related term.  The base query seeding
per-database adaptations is:

\begin{verbatim}
TITLE("federated learning") AND
ABSTRACT("federated distillation" OR
         "knowledge distillation" OR
         "logit distillation" OR
         "teacher-student" OR
         "soft labels") AND
ABSTRACT("synthetic data" OR "data-free" OR
         "deep inversion" OR diffusion OR
         GAN OR proxy)
\end{verbatim}

Each database required syntactic adaptations (field tags, Boolean
operators, quotation handling), documented in the supplementary
protocol; all queries were filtered to article- or conference-type,
English-language records.

\subsection{Eligibility criteria}
\label{sec:method_criteria}

Eligibility was defined a priori and applied consistently at both
the title/abstract and full-text stages, with each decision backed
by explicit evidence recorded in the extraction workbook.
Table~\ref{tab:eligibility_criteria} consolidates the four
\emph{inclusion criteria} (IC), all of which must be satisfied, and
the four \emph{exclusion criteria} (EC), any one of which excludes a
study.

\begin{table}[!htbp]
  \centering
  \caption{Eligibility criteria.  All four inclusion criteria (IC)
    must be satisfied; a study is excluded if any single exclusion
    criterion (EC) applies.  Evidence sources in parentheses.
    \textbf{T/A}: title and abstract.}
  \label{tab:eligibility_criteria}
  \small
  \begin{tabular}{@{}lp{0.82\linewidth}@{}}
    \toprule
    \textbf{Code} & \textbf{Criterion and operationalization} \\
    \midrule
    IC1 & \textbf{Federated learning with distillation} --- targets
          FL and replaces or augments parameter exchange with
          prediction, soft-label, or logit transfer (T/A; Methods). \\[2pt]
    IC2 & \textbf{Proxy or synthetic reference} --- uses a shared
          non-private reference (synthetic, data-free inversion,
          GAN/diffusion, or curated proxy) to compute or aggregate
          predictions (T/A; Methods). \\[2pt]
    IC3 & \textbf{Article type and language} --- peer-reviewed
          journal/conference (selected workshops) or preprint with
          code/data; English or Portuguese (Venue; PDF). \\[2pt]
    IC4 & \textbf{Replicable method description} --- sufficient
          detail to reproduce the distillation flow, the proxy
          mechanism and policy, and the evaluation protocol
          (Methods; Supp.). \\
    \midrule
    EC1 & \textbf{Non-federated or without distillation/proxy} ---
          outside the target intervention (T/A). \\[2pt]
    EC2 & \textbf{Survey, tutorial, editorial, or non-verifiable
          report} --- not a primary extraction unit; lacks replicable
          methods or experiments (Venue; PDF). \\[2pt]
    EC3 & \textbf{Insufficient methodological detail} --- distillation
          flow, proxy mechanism, or evaluation protocol below the
          minimum reporting standard (Full text). \\[2pt]
    EC4 & \textbf{Duplicate record} --- multiple versions of the same
          study; the most complete version is retained (Metadata). \\
    \bottomrule
  \end{tabular}
\end{table}

\subsection{Study selection}
\label{sec:method_selection}

\begin{figure}[!htbp]
  \centering
  \includegraphics[width=.82\linewidth,height=.38\textheight,keepaspectratio]{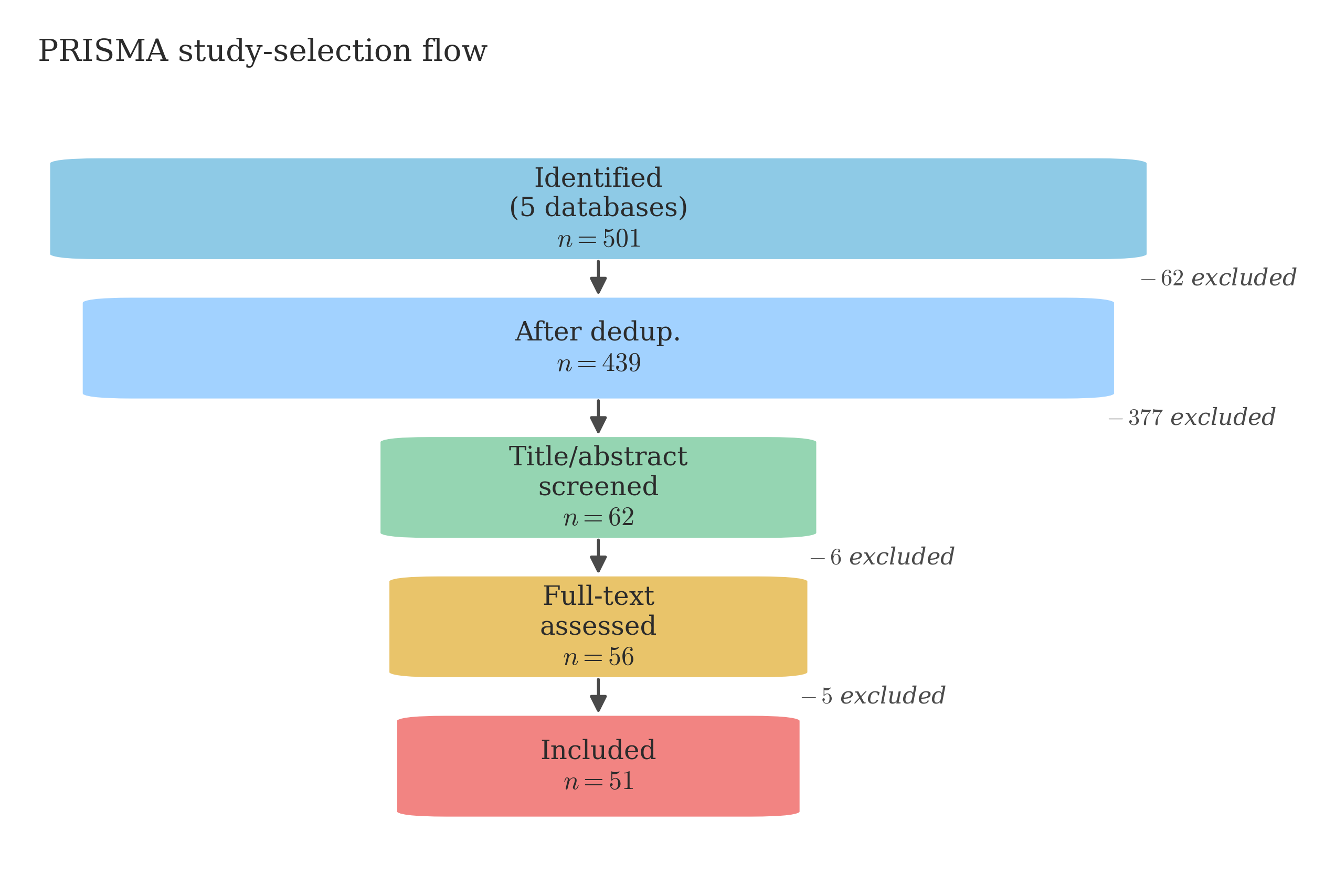}
  \Description{Funnel diagram showing record counts at each PRISMA stage, from identification to final inclusion.}
  \caption{PRISMA-style study-selection flow.  Of the 501~records
    identified across five databases, 62~duplicates were removed and
    377 of the 439 unique records were excluded at title/abstract
    screening; of the 62 candidate reports, 6 could not be retrieved
    or were ineligible at retrieval, and the 56 reports assessed at
    full text yielded 51~included primary studies (5~excluded).  The
    right margin reports the number of records excluded at each
    stage.}
  \label{fig:prisma_funnel}
\end{figure}

Selection followed four stages, summarized with exclusion counts in
Figure~\ref{fig:prisma_funnel}.  Automatic and manual deduplication
removed 62 of the 501 records; title/abstract screening against the
criteria of Section~\ref{sec:method_criteria} excluded 377 of the
439 unique records, retaining 62 candidates; full texts were sought
for all 62, with six unretrievable or ineligible at retrieval, and
the criteria plus the quality checklist were re-applied to the 56
reports assessed; backward and forward snowballing on the most
central studies yielded no additional eligible records.  A total of
\textbf{51 primary studies} was retained.  Of the five excluded at
full text, three lacked any shared proxy or reference set (EC1) and
two used synthetic data without a knowledge-distillation mechanism.
Throughout the process, LLM-based tools supported non-decisional
tasks such as summarization and metadata extraction; all eligibility
decisions were made by human reviewers, and every inclusion or
exclusion is backed by page-level evidence in the workbook.

\subsection{Quality assessment}
\label{sec:method_qa}

Each included study was assessed against an eight-item checklist
covering clarity of objectives, methodological specification,
reproducibility of the protocol, adequacy of baselines and metrics,
and artifact availability.  Scores were assigned by the first author
and subsequently reviewed in consensus meetings with two co-authors;
disagreements were resolved by discussion until unanimity was
reached.  Because the process was consensus-based rather than
conducted as independent parallel assessments, no inter-rater
agreement statistic (e.g., Cohen's~$\kappa$) is reported.
Table~\ref{tab:quality_checklist} lists the items (Q1--Q8) and their
rubric; each is scored Yes (1.0), Partial (0.5), or No (0.0) with
default weight 1.0, so the total is

\begin{equation}
  \label{eq:qa_total}
  S \;=\; \sum_{i=1}^{8} w_i \, s_i, \qquad S \in [0, 8].
\end{equation}

Quality is not used as an exclusion criterion; it enables
sensitivity analyses and surfaces methodological gaps
(Section~\ref{sec:results}).  Across the 51 studies the mean total
is $\bar{S} = 6.70$ (median 7.00, standard deviation 0.83, range
$[4.0, 8.0]$), indicating generally adequate reporting, with the
lowest scores concentrated in Q8 (reproducibility artifacts).  The
full per-item log, with scores, weights, and supporting evidence, is
stored in the extraction workbook.

\begin{table}[!htbp]
  \centering
  \caption{Quality checklist items, scoring rubric, and default
    weights.  Each item is scored as \textbf{Y}es (1.0), \textbf{P}artial
    (0.5), or \textbf{N}o (0.0); $W$ denotes the default weight.}
  \label{tab:quality_checklist}
  \small
  \begin{tabular}{@{}lp{6cm}p{3.8cm}c@{}}
    \toprule
    \textbf{ID} & \textbf{Question (minimum reporting standard)} & \textbf{Scoring rubric} & \textbf{W} \\
    \midrule
    Q1 & Domain, application, and non-IID scenario clearly described.
       & Y: domain + split explicit; P: partial; N: absent. & 1.0 \\[2pt]
    Q2 & Objectives or hypotheses about proxy/synthetic use stated.
       & Y: explicit; P: generic aim; N: absent. & 1.0 \\[2pt]
    Q3 & Distillation flow replicable (roles, predictions, aggregation).
       & Y: full flow with symbols; P: partial; N: absent. & 1.0 \\[2pt]
    Q4 & Proxy mechanism detailed (generator, curation, sampling).
       & Y: mechanism and policy; P: mechanism only; N: absent. & 1.0 \\[2pt]
    Q5 & Proxy-quality indicators linked to outcomes
         (diversity, realism, coverage).
       & Y: indicators with analysis; P: indicators only; N: absent. & 1.0 \\[2pt]
    Q6 & Experimental protocol fair (non-IID setup, baselines, seeds).
       & Y: all specified; P: partial; N: absent. & 1.0 \\[2pt]
    Q7 & Robustness analysis (noise, poisoning, OOD) and limitations.
       & Y: threats, results, and limits; P: partial; N: absent. & 1.0 \\[2pt]
    Q8 & Reproducibility artifacts (code, data, seeds) released or
         documented.
       & Y: artifacts or full specification; P: partial; N: absent. & 1.0 \\
    \bottomrule
  \end{tabular}
\end{table}

\subsection{Data extraction and synthesis}
\label{sec:method_extraction}

Extraction used a spreadsheet workbook: a \texttt{Study\_Register}
sheet records shared bibliographic and contextual fields for every
study (identifier, title, year, venue, type, DOI, eligibility,
reviewer, extraction date, domain and task, dataset summary, FL
setting, heterogeneity type, client count, and code/data
availability); six additional sheets, one per research question,
record the RQ-specific fields consolidated in Supplementary
Table~S1; and a \texttt{Controlled\_Vocabulary} sheet enforces
consistent terminology across categorical fields.  Every extracted
field carries an explicit \texttt{KeyEvidence} entry pointing to the
supporting page, figure, or table, and an \texttt{AnalystNote} for
qualitative observations.  Synthesis proceeds in two complementary
modes, following the spirit of narrative synthesis in
software-engineering reviews~\cite{kitchenham2007guidelines}: a
\emph{descriptive} characterization of the corpus along the
\texttt{Study\_Register} dimensions
(Section~\ref{sec:results_overview}), and a \emph{thematic}
synthesis that groups evidence by the three taxonomy axes, uses
cross-tabulations of controlled-vocabulary fields to surface
structural patterns, and cross-walks proxy-quality indicators (RQ4)
against the outcomes of RQ3.

\subsection{Threats to validity}
\label{sec:method_threats}

\textbf{Construct validity.}  The target intersection is
terminologically fragmented (``proxy,'' ``transfer set,''
``reference data,'' ``pseudo-data,'' ``distillation buffer'').  The
search string combines distillation- and synthetic-proxy-related
term families with Boolean AND, and extraction uses a controlled
vocabulary that harmonizes surface variants; borderline cases (e.g.,
dataset distillation without teacher--student transfer) required
full-text adjudication against EC1/EC3.

\textbf{Internal validity.}  Selection and extraction involve
judgement.  Reviewer bias was reduced by making all decisions
traceable to explicit evidence, applying a predefined checklist, and
reserving LLM-based tools for non-decisional tasks; the selection
results were cross-checked against the PRISMA flow, and
discrepancies were resolved by re-reading the full text and updating
the workbook.

\textbf{External validity.}  The review is restricted to five
databases and the 2019--2025 window; regional or specialized venues
may be underrepresented, and the deliberate exclusion of preprints
means the most recent unrefereed contributions are not
captured---given the field's velocity, a real cost, accepted in
exchange for the quality assurance of peer review.  The strong
temporal concentration in 2024--2025
(Section~\ref{sec:results_overview}) suggests that most relevant
literature falls within the window.

\textbf{Conclusion validity.}  The synthesis combines descriptive
statistics with thematic analysis and does not attempt meta-analytic
pooling, which the heterogeneity of tasks, baselines, and metrics
would render inappropriate; conclusions are framed as patterns, and
the workbook provides full traceability from every synthesis
statement back to the primary studies.

\section{Results and Thematic Synthesis}
\label{sec:results}

This section organizes the evidence from the 51 included studies
around the six research questions of Section~\ref{sec:method_rqs}:
descriptive overview (Section~\ref{sec:results_overview}), proxy use
(RQ1), generation mechanisms (RQ2), effects on communication,
privacy, and performance (RQ3), proxy quality (RQ4), threats and
defenses (RQ5), and methodological gaps (RQ6).

\subsection{Descriptive overview of included studies}
\label{sec:results_overview}

The final corpus comprises 51 primary studies published between 2020
and 2025 across 33~journals, 17~conferences, and 1~preprint.

\begin{figure}[!htbp]
  \centering
  \includegraphics[width=.82\linewidth,height=.42\textheight,keepaspectratio]{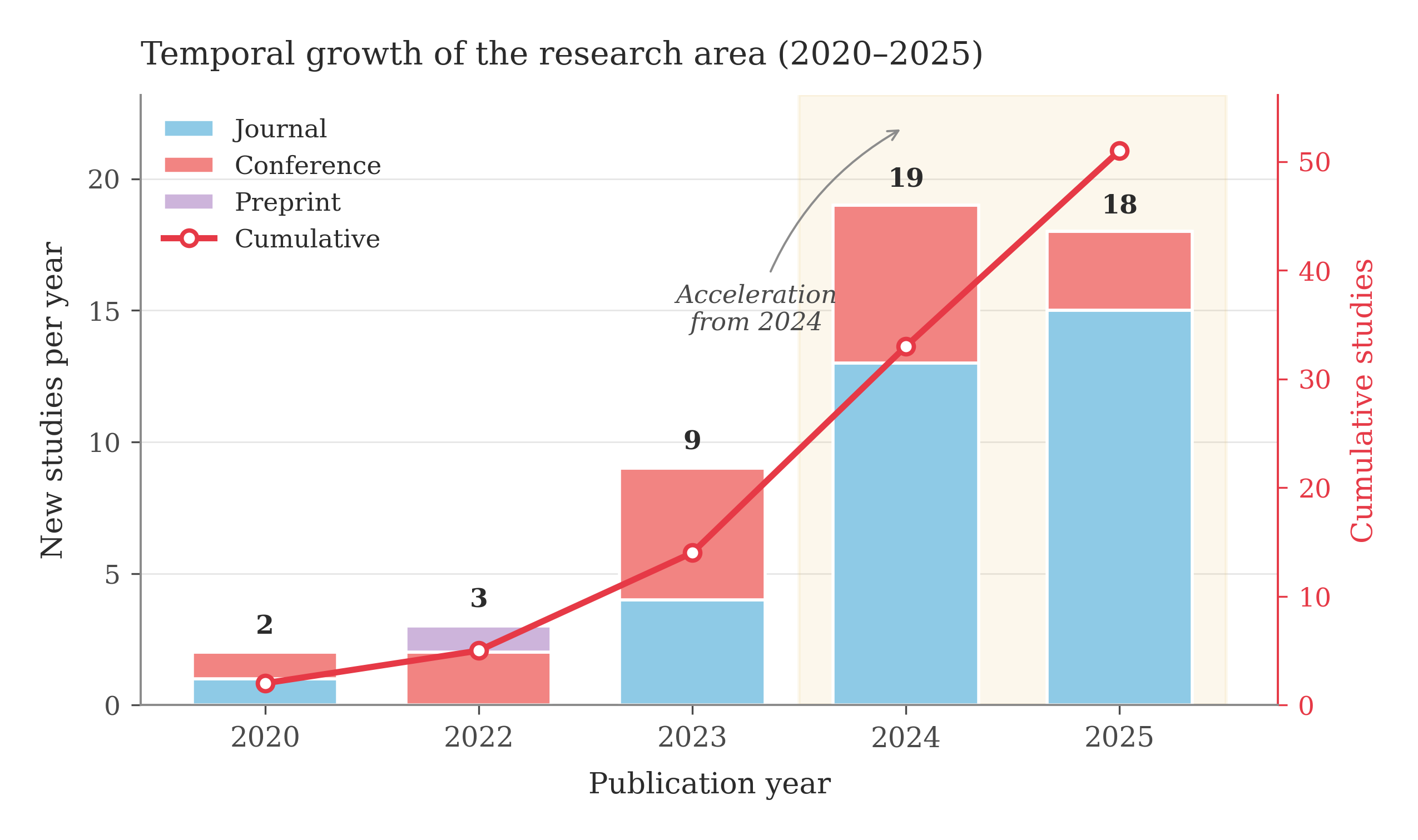}
  \Description{Bar chart of publications per year with an overlaid cumulative growth curve.}
  \caption{Temporal growth of the research area (2020--2025).  Stacked
    bars show the number of new studies per year by publication type
    (journal, conference, preprint); the red line and right-axis
    report the cumulative number of studies.  The shaded region
    highlights the acceleration from 2024 onward.  Of the 51~studies,
    37 (73\,\%) were published in 2024 or 2025.}
  \label{fig:growth_cumulative}
\end{figure}

\textbf{Temporal distribution.}
Figure~\ref{fig:growth_cumulative} reports the number of studies per
year.  The field is very young: the earliest studies appear in 2020
($n = 2$), production remains modest through 2022 ($n \le 3$ per
year), a clear acceleration takes place in 2023 ($n = 9$), and
2024--2025 jointly account for 37~studies---73\,\% of the entire
corpus.  The publication-type mix also shifts: conferences dominate
before 2023, after which journal articles overtake them at roughly
2:1, reaching 65\,\% of the corpus overall (33 of 51) against 33\,\%
conferences; a single preprint was retained under IC3 for its
substantive contribution and released artifacts.  This concentration
indicates that the intersection of federated distillation with
synthetic proxies has crystallized as a distinct topic only in the
last two years.

\textbf{Application domains.}
Computer vision dominates, appearing in 24~studies (47\,\%),
followed by healthcare with 14 (27\,\%)---clinical and
medical-imaging applications whose regulatory constraints (HIPAA,
GDPR, LGPD) make distillation an attractive alternative to
parameter-based FL.  Industrial applications (9), cybersecurity (8),
IoT and human-activity recognition (7), graph and social-media
analysis (6), edge/networks (5), NLP and speech (3), energy (2), and
recommender systems and reinforcement learning (1 each) complete the
picture, with a residual ``Other'' long tail (19~studies, 37\,\%) of
specialized applications (Supplementary Fig.~S2).  This
concentration means that advances in vision and healthcare set the
tone for the field, but it also raises an external-validity
question---whether methods validated on image datasets transfer to
tabular, time-series, or text-heavy domains---to which we return in
Section~\ref{sec:results_rq6}.

\textbf{Federated setting and heterogeneity.}
Cross-device FL dominates with 34~studies (67\,\%); cross-silo
appears in 11 (22\,\%), mostly healthcare and industrial; hybrid
setups and unreported settings account for 3~studies (6\,\%) each
(Supplementary Fig.~S3).  Mixed heterogeneity is the default:
38~studies (75\,\%) combine multiple non-IID axes, most commonly
label plus quantity skew generated via Dirichlet partitioning of the
training data~\cite{hsu2019measuring} with concentration parameter
$\alpha$ typically sampled from $\{0.1, 0.5, 1.0\}$; label-skew-only
setups appear in 8~studies (16\,\%), feature skew in 3, IID in 1,
and 1 study does not report the configuration.  The pairings are not
arbitrary: cross-device studies concentrate in the mixed cell
($n = 29$), while cross-silo studies are spread more evenly and
contribute all three feature-skew evaluations and the single IID
baseline---a direct reflection of institutional data (different
imaging protocols across hospitals) versus device data (labels and
quantities diverge, features less so).

\textbf{Experimental scale.}
The client count is reported in 48 of 51 studies (94\,\%).  The
median is 10~clients (IQR 6--22, range 2--179); 28~studies (58\,\%
of those reporting) use 10 or fewer, and only 5 scale beyond 50
(Supplementary Fig.~S4).  This falls far short of the $10^3$--$10^6$
populations envisioned for cross-device FL~\cite{kairouz2021advances},
so reported results may \emph{overestimate} effectiveness at
realistic scale, where aggregating many heterogeneous teachers and
covering the union of local distributions both become harder.

\textbf{Benchmark datasets.}
The corpus concentrates on small image-classification benchmarks:
MNIST (17~studies, 33\,\%), CIFAR-10 (14, 27\,\%), CIFAR-100 (9,
18\,\%), Fashion-MNIST (8, 16\,\%), and SVHN (5, 10\,\%), with
ImageNet-scale data in only 1--2 studies and domain-specific
datasets (ChestX-ray14, NSL-KDD, CWRU) appearing once or twice each
(Supplementary Fig.~S5).  Since MNIST-family benchmarks tolerate
weak teachers, methods not validated beyond CIFAR should be
interpreted with caution.

\textbf{Summary.}
Four facts frame the thematic synthesis: the field is recent and
growing rapidly; computer vision and healthcare dominate the
application landscape; cross-device FL with mixed heterogeneity is
the modal configuration, with cross-silo studies contributing the
structurally richer heterogeneity evaluations; and the experimental
scale is modest, raising external-validity concerns.

\subsection{RQ1: Use of synthetic/generated data as proxies}
\label{sec:results_rq1}

\subsubsection{Proxy types}
\label{sec:rq1_types}

\begin{figure}[!htbp]
  \centering
  \includegraphics[width=.75\linewidth,height=.38\textheight,keepaspectratio]{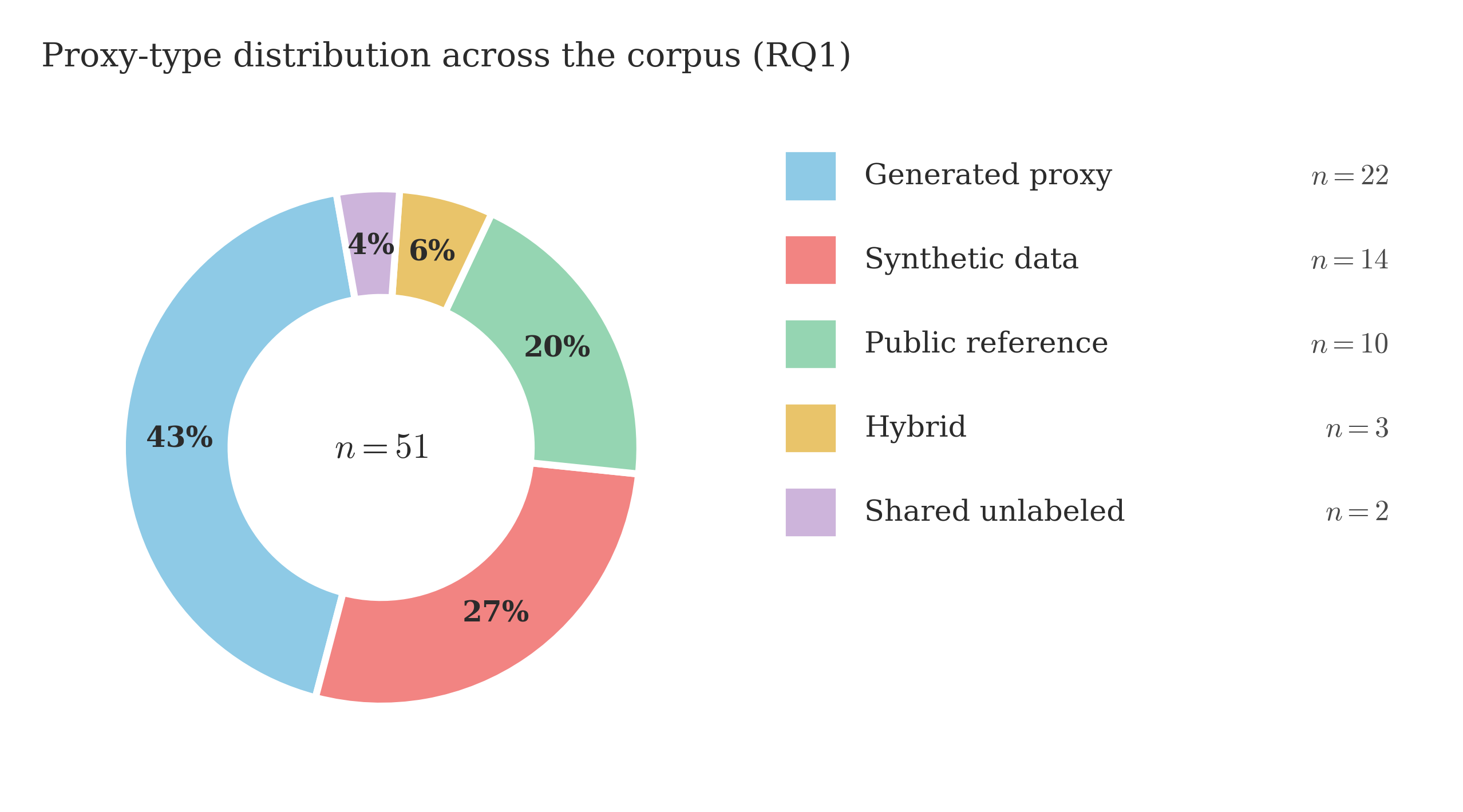}
  \Description{Donut chart of the five proxy-type categories with their percentage shares.}
  \caption{Distribution of proxy types across the 51 included
    studies.  Generated proxies (GAN, diffusion, VAE, deep
    inversion) account for 43\,\% of the corpus, followed by
    synthetic non-generative data (27\,\%) and public references
    (20\,\%).  Hybrid and shared-unlabeled schemes together account
    for the remaining 10\,\%.}
  \label{fig:proxy_type_donut}
\end{figure}

Figure~\ref{fig:proxy_type_donut} shows the distribution of proxy
types, harmonized into five categories.  \textbf{Generated proxies}
(22~studies, 43\,\%)---samples produced by a GAN, diffusion model,
VAE, or deep inversion trained within or by the federation---are
exemplified by Fed2KD's two-way distillation for pandemic risk
assessment~\cite{sun2022fed2kd}, the recommender framework
FedDyn~\cite{jin2023feddyn}, GAN-based optimization for vehicular
task offloading in 6G~\cite{wu2024efficient}, diffusion-based
pseudo-data for non-IID FedAvg~\cite{najafi2024enhancing} and
multi-center medical segmentation~\cite{zhang2025non}, and
pseudo-graph generation in FedPPD~\cite{lin2025fedppd}.
\textbf{Synthetic (non-generative) data} (14~studies,
27\,\%)---augmentation, perturbation, statistical replay, or dataset
distillation---includes the clustering-based construction of
FedCK~\cite{chen2025fedck}, meta-data-driven
extraction~\cite{yin2025meta}, privacy-preserving distillation for
industrial cyber-physical systems~\cite{liang2023privacy,
liang2024enhanced}, data-free filtering~\cite{lu2024data},
PeFAD for time-series anomaly detection~\cite{xu2024pefad}, and the
one-shot synthetic distillates of FedSD2C~\cite{zhang2024one}.
\textbf{Public references} (10~studies, 20\,\%) continue the
FedMD/FedDF paradigm~\cite{li2019fedmd, lin2020ensemble}:
layer-wise cross-device distillation~\cite{le2023layer} and its
follow-up CDKT-FL~\cite{le2024cdkt}, the fault-diagnosis framework
HFTL-KD~\cite{lu2024hftl}, vertical federated knowledge transfer in
healthcare~\cite{huang2023vertical},
FedKD-CPI for compound--protein interaction~\cite{wang2025fedkd},
and the oldest included study, FedReID~\cite{zhuang2020performance}.
\textbf{Hybrid} schemes (3~studies, 6\,\%) combine
mechanisms~\cite{jin2023feddyn, chen2023privacy,
chadha2024training}, and \textbf{shared unlabeled} pools (2~studies,
4\,\%) support cross-client alignment via knowledge
filtering~\cite{lu2025fedfaa} and multimodal embedding
transfer~\cite{le2025fedmekt}.

The combined share of generated-plus-synthetic-plus-hybrid proxies
reaches 77\,\% of the corpus (39 of 51), confirming that the field
is moving decisively away from the assumption of a pre-existing
suitable public dataset.

\subsubsection{Distillation objects and flow directions}
\label{sec:rq1_flows}

\textbf{Distillation object.}  Of the 51 studies, 27 communicate
\emph{logit vectors}, 13 \emph{softened class probabilities}
(Eq.~\eqref{eq:softmax_temperature}), 8 \emph{intermediate features}, and 3
other forms (prototypes, attention maps, representation pairs).
Logits and soft labels together account for 78\,\%, confirming the
dominance of response-based distillation~\cite{hinton2015distilling}.
Feature-based distillation requires compatible intermediate
representations and is confined to controlled cross-silo settings
such as heterogeneous fault diagnosis~\cite{wang2024towards} and
graph-augmented fMRI analysis~\cite{wang2025graph}.

\begin{figure}[!htbp]
  \centering
  \includegraphics[width=.82\linewidth,height=.42\textheight,keepaspectratio]{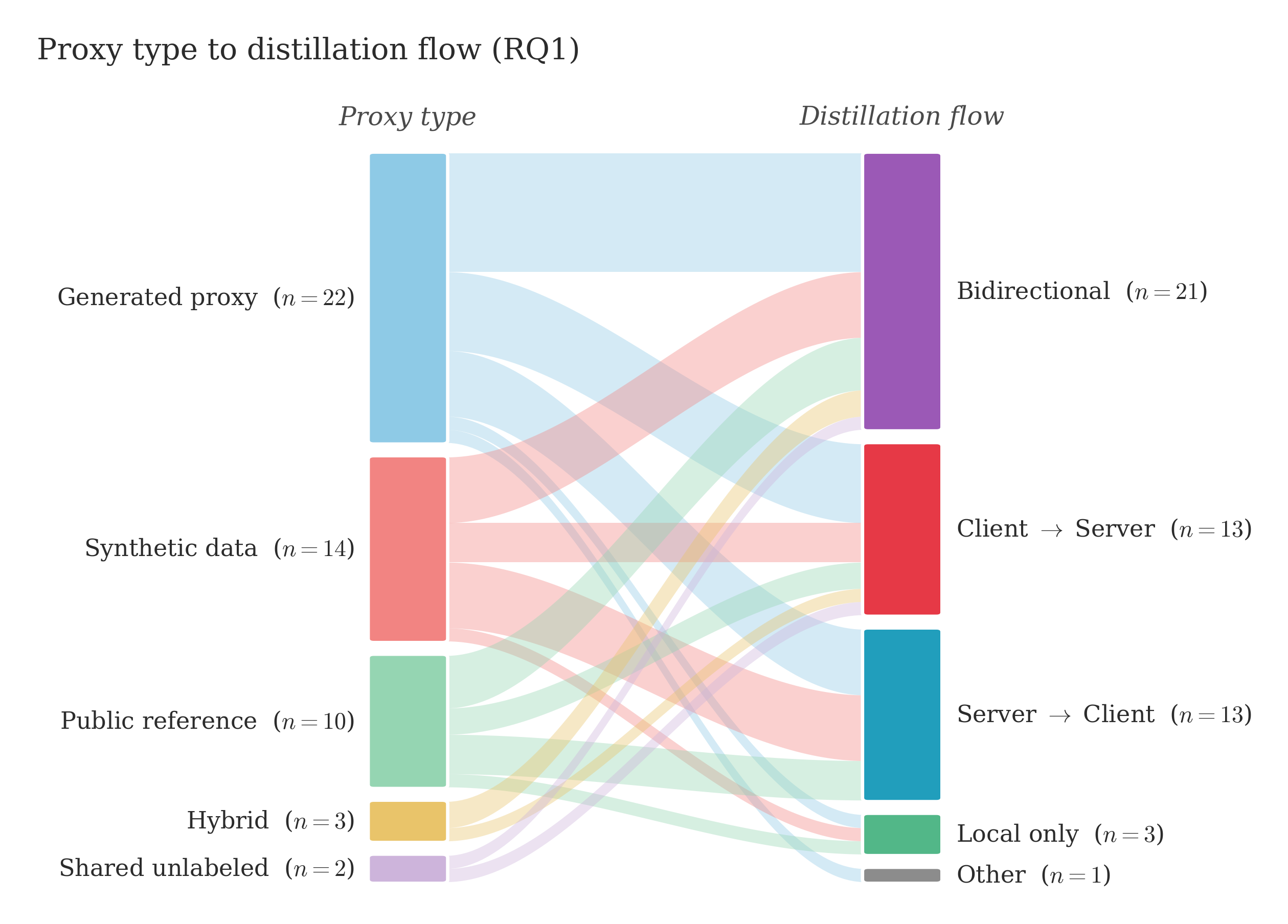}
  \Description{Alluvial diagram linking proxy types to distillation flows and ownership scopes.}
  \caption{Alluvial diagram showing how proxy types map onto
    distillation flows across the corpus.  Band widths are
    proportional to the number of studies.  Bidirectional flow is
    the modal choice (41\,\% of the corpus); client-to-server and
    server-to-client flows are balanced at 26\,\% each; local-only
    distillation is rare.  Generated proxies are the primary source
    of bidirectional schemes.}
  \label{fig:proxy_to_flow_sankey}
\end{figure}

\textbf{Flow direction.}  Figure~\ref{fig:proxy_to_flow_sankey} maps
proxy types onto distillation flows.  \emph{Bidirectional} flow is
the modal choice (21~studies, 41\,\%), with predictions exchanged in
both directions within the same round, as in
FedTweet~\cite{wang2024fedtweet}, Fed2KD~\cite{sun2022fed2kd}, and
FedACK~\cite{yang2023fedack}.  \emph{Client-to-server} and
\emph{server-to-client} flows are balanced at 13~studies each
(26\,\%): the former is the classical pattern of
FedCK~\cite{chen2025fedck}, FedFAA~\cite{lu2025fedfaa},
FedDE~\cite{hao2024fedde}, and flow-based intrusion
detection~\cite{li2025flow}; the latter presupposes a useful
server-side teacher, as in Fedadkd~\cite{song2024fedadkd},
data-independent knowledge transfer~\cite{zhang2023towards},
FedCMD~\cite{bano2024fedcmd}, and one-shot medical
FL~\cite{kang2025efficient}.  \emph{Local-only} distillation
(3~studies, 6\,\%) confines soft targets to the client, as in
FedKG~\cite{wang2022fedkg} and personalized long-tailed
distillation~\cite{lv2024personalized}.  The prevalence of
bidirectional plus client-to-server flow (67\,\% together) indicates
that most of the corpus relies on clients as the teachers of the
system.

\subsubsection{Proxy ownership scope}
\label{sec:rq1_ownership}

Five ownership scopes recur: \textbf{globally shared} (19~studies,
37\,\%), \textbf{server-side} (16, 31\,\%), \textbf{hybrid} (10,
20\,\%), \textbf{client-side} (5, 10\,\%), and
\textbf{group-specific} (1, 2\,\%).  The concentration in the first
two categories (together 68\,\%) reflects the centralized-server
architecture that dominates the field; the client-side studies are
recent and explore on-device synthesis that eliminates the need to
transmit a common reference---promising under severe bandwidth
constraints, but raising new coordination questions.

\subsubsection{Cross-sectional patterns across the three axes}
\label{sec:rq1_patterns}

The three design axes are orthogonal in principle but strongly
correlated in practice (Supplementary Fig.~S6).  Three structural
patterns emerge.  \textbf{Pattern 1: generated proxies enable richer
flows.}  Generated proxies spread relatively uniformly across flow
directions (9 bidirectional, 6 client-to-server, 5 server-to-client,
1 local, and 1 other), whereas public references concentrate in bidirectional (4)
and server-to-client (3) flows, avoiding the client-as-teacher
position.  \textbf{Pattern 2: ownership is tightly coupled to proxy
type.}  Public references are overwhelmingly globally shared (9 of
10); generated proxies are predominantly server-side (11 of 22) or
hybrid (5 of 22); synthetic proxies span the full ownership
spectrum---from server-side to client-side---and include the
corpus's only group-specific case.
\textbf{Pattern 3: bidirectional flow pairs with globally
visible proxies.}  Of the 21 bidirectional studies, 18 use globally
shared (11) or hybrid (7) ownership, while purely server-side (1)
and client-side (2) ownership are rare: exchanging predictions in
both directions presupposes a reference that both ends can evaluate,
which favors proxies that are globally visible or jointly
maintained.  The effective design space is therefore
much smaller than the full cross-product of the axes would suggest.

\subsubsection{Summary of RQ1}

Three findings summarize RQ1.  First, 77\,\% of the included studies
use generated, synthetic, or hybrid proxies, and only 20\,\% adopt a
purely public reference.  Second, logits and soft labels remain the
dominant distillation objects (78\,\%), with feature- and
prototype-based transfer confined to specialized settings.  Third,
the combinations of proxy type, flow, and ownership define a small
number of recurring archetypes---\emph{generated-server-%
bidirectional}, \emph{public-global-unidirectional}, and
\emph{synthetic-client-local}---that recur throughout the remainder
of this synthesis.

\subsection{RQ2: Proxy generation and selection mechanisms}
\label{sec:results_rq2}

\subsubsection{Mechanism families and temporal evolution}
\label{sec:rq2_families}

\begin{figure}[!htbp]
  \centering
  \includegraphics[width=.82\linewidth,height=.42\textheight,keepaspectratio]{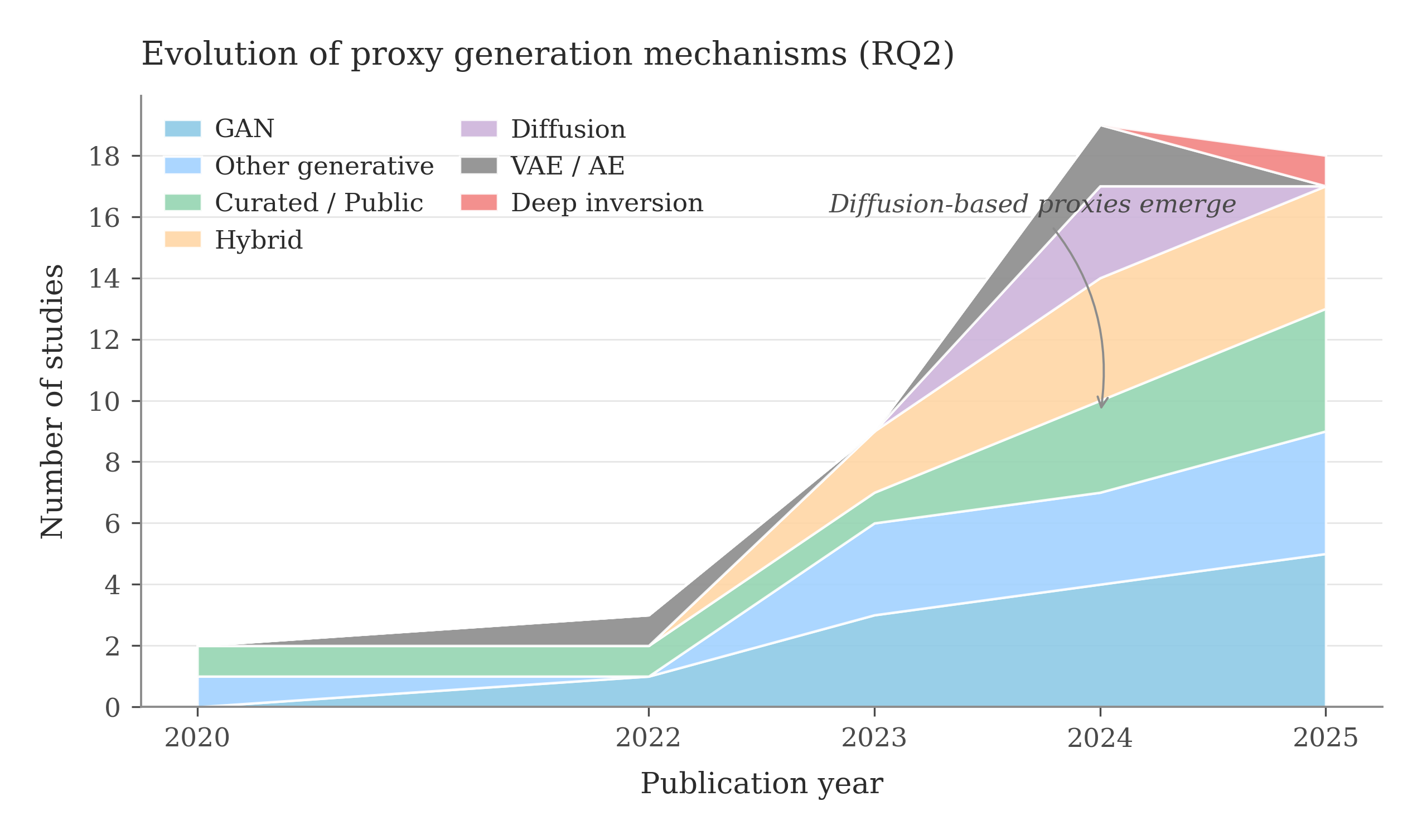}
  \Description{Stacked area chart of proxy mechanism families across publication years.}
  \caption{Evolution of proxy generation and selection mechanisms
    across the corpus, 2020--2025 (stacked area chart).  GAN-based
    generation has been the modal mechanism since 2023; hybrid and
    ``other generative'' families grew most rapidly in 2024--2025;
    diffusion-based proxies emerged in 2024 and remain rare.  Deep
    inversion appears only in 2025 and is currently the smallest
    family.}
  \label{fig:mechanism_timeline}
\end{figure}

Figure~\ref{fig:mechanism_timeline} shows the evolution of mechanism
families.  \textbf{GAN-based generation} is the modal family
(13~studies, 25\,\%): conditional GANs enable class-balanced
synthesis under label skew~\cite{chen2025fedck, liu2025rapid,
yi2025fedfld}, ACGAN variants add label prediction to the
discriminator~\cite{chen2023privacy}, and Wasserstein training with
gradient penalty stabilizes industrial deployments~\cite{
liang2023privacy, liang2024enhanced}.  The breadth of GAN usage
spans backdoor-aware sample generation in ADFL~\cite{zhu2023adfl},
group-wise local GANs~\cite{yu2025heterogeneous}, semi-supervised
per-machine GANs for fault diagnosis~\cite{karami2024gfd}, and even
procedurally rendered Blender imagery~\cite{maron2024synthetic}.
\textbf{Curated/public proxies} (10~studies, 20\,\%) and
\textbf{hybrid schemes} (10~studies, 20\,\%) follow; the most common
hybrid pattern is server-side generation conditioned on aggregated
client information~\cite{fu2024federated, lin2025fedppd,
lv2024personalized, qu2025darkdistill, wang2024fedtweet,
zheng2024fedgkd}, with FedDyn coupling generated candidates to
top-$K$ confidence selection~\cite{jin2023feddyn} and PSI-based
sample identification in vertical FL~\cite{huang2023vertical}; 6 of
the 10 hybrid studies date from 2024--2025.  An \textbf{other
generative} bucket (11~studies, 22\,\%) collects label- and
noise-conditioned lightweight generators with custom
objectives~\cite{li2025flow, lu2024data, yang2023fedack,
zhang2023towards, zhao2024federated}, generative mapping for fault
diagnosis~\cite{wang2024towards}, mask generators with cascaded
diffusion~\cite{zhang2025non}, and meta-data
synthesis~\cite{yin2025meta}.  \textbf{Diffusion-based generation}
appears in only 3~studies, all from 2024~\cite{song2024fedadkd,
najafi2024enhancing, hao2024fedde}, its adoption slowed by per-sample
sampling cost; \textbf{VAE/autoencoder} schemes also number
3~\cite{sun2022fed2kd, zhang2024one, xu2024pefad}, concentrated in
tabular and time-series domains; and \textbf{deep inversion} appears
in a single study~\cite{you2025adaptive}, despite requiring neither
real data nor a generator---its cost per synthesized sample and
diversity limitations explain the scarcity.  Temporally, GAN-based
and hybrid generation expand fastest in 2024--2025 (17 of the
37 studies of those years), while diffusion and inversion emerge
only recently and coexist with, rather than replace, the older
families.

\subsubsection{Source and curation policies}
\label{sec:rq2_policies}

Five proxy sources recur: server-generated (15~studies, 29\,\%),
public dataset (10), hybrid (10), client-generated (9, 18\,\%), and
external sources (3 plus minor variants).  Server generation mirrors
the centralized architecture; the growing client-generated share
signals interest in on-device synthesis.  Once sourced, most proxies
undergo curation via three recurring policies:
\emph{class-balanced sampling} of labels from a target distribution
$P(y)$---uniform to compensate label skew, or the aggregated
client mixture~\cite{chen2025fedck, najafi2024enhancing};
\emph{confidence-based filtering}, retaining samples by top-$K$
interaction probability~\cite{jin2023feddyn}, ensemble confidence
scores~\cite{hao2024fedde}, or pre-aggregation knowledge
filtering~\cite{lu2025fedfaa}; and \emph{core-set or
informativeness-based selection}, via $V$-information at two
granularities~\cite{zhang2024one}, intersection of client class
boundaries~\cite{lu2024data}, or semi-supervised filtering that
preserves label-rare regions~\cite{karami2024gfd}.
Hyperparameter reporting is asymmetric: 27~studies (53\,\%) fully
report the distillation ($\tau$, $\alpha$), proxy-size ($M$), and
generator families; 16 (31\,\%) report only the distillation side;
and 8 (16\,\%) provide partial or no generator information---a
contributor to the reproducibility gap of
Section~\ref{sec:rq6_reproducibility}.

\subsubsection{Reported advantages and drawbacks}
\label{sec:rq2_tradeoffs}

Five themes recur.  On the advantage side:
(1)~\emph{independence from suitable public datasets}, the most
cited benefit, especially in regulated
domains~\cite{chen2025fedck, yi2025fedfld, karami2024gfd};
(2)~\emph{balance under label skew} via class-conditional
generation, improving convergence and accuracy under skewed
Dirichlet partitions~\cite{hsu2019measuring} with
$\alpha < 0.5$~\cite{liang2023privacy, liu2025rapid}; and
(3)~\emph{privacy preservation beyond raw data}, transferring
synthetic samples or soft predictions instead of
gradients~\cite{liang2024enhanced, chen2023privacy, yin2025meta}.
On the drawback side: (4)~\emph{computational overhead}---iterative
diffusion sampling increases server-side training
time~\cite{najafi2024enhancing}, one-shot distillates add local
overhead~\cite{zhang2024one}, and returns diminish as the proxy
size $M$ grows~\cite{zhang2023towards}; and (5)~\emph{stability and
hyperparameter sensitivity}---CVAEs may fail to converge under weak
regularization~\cite{sun2022fed2kd}, prior generative methods show
overfitting risk~\cite{song2024fedadkd}, and generator outputs
drift across rounds~\cite{yi2025fedfld}.  A subtler residual risk is
that synthetic proxies are not intrinsically private: membership
inference and model inversion against generator outputs or soft
labels remain unaddressed by most frameworks~\cite{chen2025fedck},
motivating the differential-privacy layering examined in
Section~\ref{sec:rq3_privacy}.

\subsubsection{Summary of RQ2}

Four findings summarize RQ2.  First, five families account for all
included methods, with GANs as the modal choice (25\,\%).  Second,
diffusion and deep inversion are the most recent and rarest
families, despite their centralized-synthesis pedigree.  Third,
server-side generation is the modal source (29\,\%), but
client-generated and hybrid proxies jointly reach 37\,\% and are
growing.  Fourth, the reported advantages (dataset independence,
label-skew mitigation, privacy) are mirrored by recurring drawbacks
(overhead, instability, residual privacy risk).

\subsection{RQ3: Effects on communication, privacy, and performance}
\label{sec:results_rq3}

\subsubsection{Non-functional reporting practices}
\label{sec:rq3_reporting}

\begin{figure}[!htbp]
  \centering
  \includegraphics[width=.82\linewidth,height=.42\textheight,keepaspectratio]{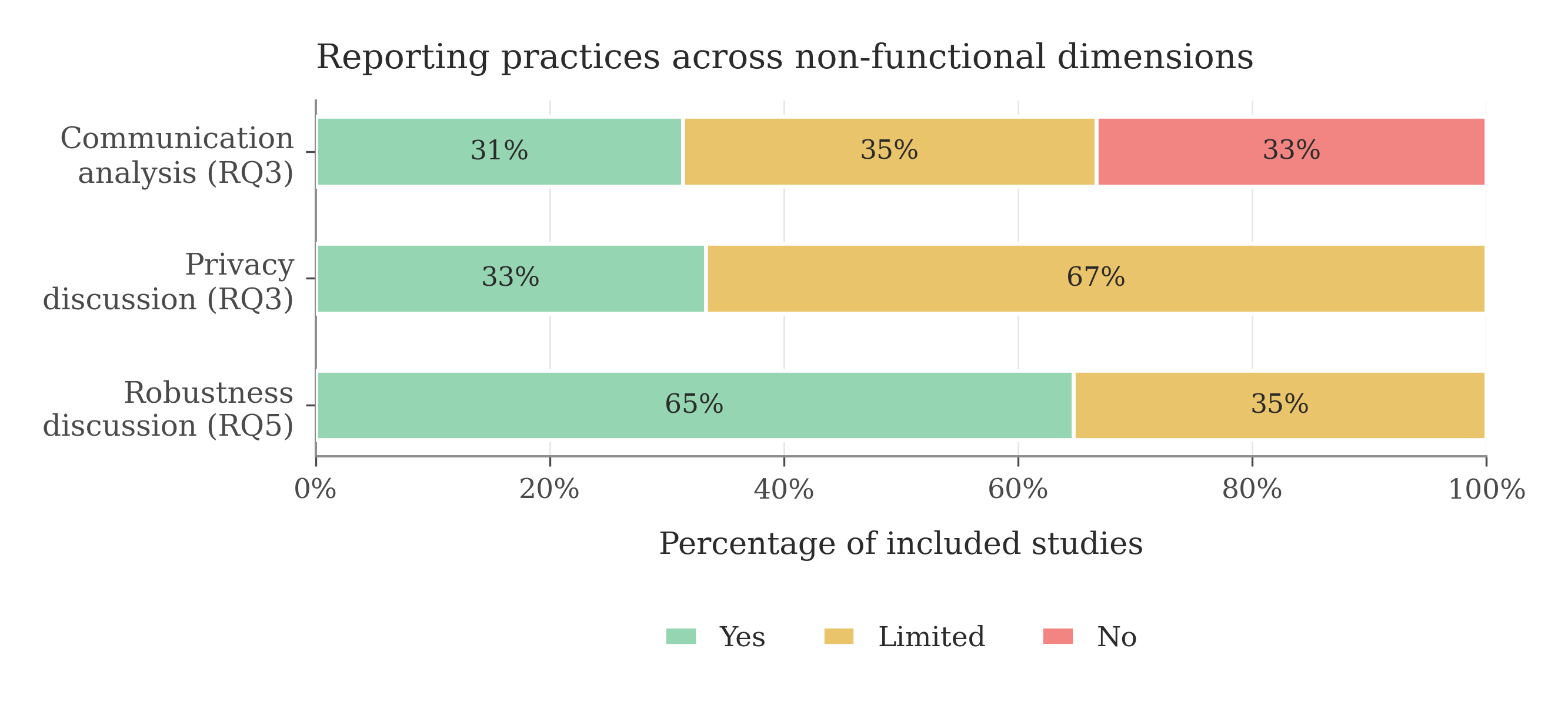}
  \Description{Heat strip indicating, for each study, which non-functional dimensions are reported.}
  \caption{Reporting practices across non-functional dimensions.
    For each of communication analysis (RQ3), privacy discussion
    (RQ3), and robustness discussion (RQ5), the bars show the
    fraction of the 51 included studies that report the dimension
    with substantive depth (Yes), in passing (Limited), or not at
    all (No).  Substantive reporting is the minority position
    across all three dimensions.}
  \label{fig:reporting_heatstrip}
\end{figure}

Figure~\ref{fig:reporting_heatstrip} summarizes how systematically
the corpus reports its non-functional dimensions.
\textbf{Communication analysis} is substantively reported in only
16~studies (31\,\%), treated in passing in 18 (35\,\%), and absent
in 17 (33\,\%)---striking, given that communication efficiency is
one of the three motivations for federated distillation.
\textbf{Privacy} receives explicit analysis (assumption, threat
model, or defense with stated $\epsilon$) in 17~studies (33\,\%);
the remaining 34 (67\,\%) mention it only generically, though no
study omits it entirely.  \textbf{Robustness} is the weakest
dimension and is deferred to Section~\ref{sec:results_rq5}.  These
asymmetries mean our quantitative claims can only summarize what
studies \emph{say}---they cannot support meta-analytic pooling---and
the gap between stated and measured importance is itself a finding.

\subsubsection{Performance under non-IID data}
\label{sec:rq3_performance}

All 51 studies report gains over parameter-averaging
baselines---most commonly FedAvg~\cite{mcmahan2017fedavg},
FedProx~\cite{li2020fedprox}, FedDistill~\cite{jeong2018communication},
FedDF~\cite{lin2020ensemble}, and FedGen~\cite{zhu2021datafree}.
\textbf{Gain magnitude scales with heterogeneity intensity}: at
Dirichlet~\cite{hsu2019measuring} $\alpha = 0.05$ on MNIST, Fed2KD
reports up to 30\,\% accuracy improvement over FedAvg (18\,\% on
Fashion-MNIST)~\cite{sun2022fed2kd}; at $\alpha = 0.1$, FedDiff
reaches target accuracy with 57--71\,\% fewer rounds on
CIFAR-10/100~\cite{najafi2024enhancing}; and FedAdKD reaches
74.10\,\% on EMNIST at $\alpha = 0.01$ against 65.36\,\% for the
best baseline~\cite{song2024fedadkd}.  \textbf{One-shot designs
trade rounds for accuracy}: FedSD2C achieves 26.83 versus 14.35 for
the best one-shot baseline on Tiny-ImageNet at
$\alpha = 0.1$~\cite{zhang2024one, kang2025efficient}.
\textbf{Specialized domains report large gains}, partly because
baselines were not designed for the modality: compound--protein
interaction~\cite{wang2025fedkd}, bearing-fault
diagnosis~\cite{lu2024hftl, wang2024towards}, and intrusion
detection at 96.56\,\% accuracy on USTC-TFC2016 with
$\alpha = 0.1$~\cite{li2025flow}.  The dominant metric is top-1
accuracy (43~studies, 84\,\%), with precision/recall/F1 in 12
(24\,\%), plus HR@$K$/NDCG@$K$ for recommendation~\cite{jin2023feddyn}
and RMSE for prognostics~\cite{ding2024global}; convergence metrics
appear mainly alongside the communication claims below.

\subsubsection{Communication cost: how much, measured how}
\label{sec:rq3_comm}

\textbf{Rounds is the modal metric} (12 of the 16 substantive
analyses): 57--71\,\% round reductions~\cite{najafi2024enhancing},
consistent round savings~\cite{lu2024data}, and reduced rounds in
balanced configurations~\cite{hao2024fedde}---but rounds hide the
per-round payload.  \textbf{Byte-based reporting remains rare}
(9~studies, 18\,\%): FedSD2C reports $\approx$4\,\% of the
parameter-sharing communication cost in one shot~\cite{zhang2024one};
PeFAD reports 1{,}800\,GB versus 3{,}060\,GB for full fine-tuning on
SMD~\cite{xu2024pefad}; FedCMD details 8.05--60.37\,MB per round
across client counts~\cite{bano2024fedcmd}; GFD reports cumulative
kilobytes per round~\cite{karami2024gfd}; and the earliest byte-level
study reports payload size and communication
period~\cite{cha2020proxy}.  \textbf{Idiosyncratic metrics fragment
the literature}---transmitted parameter capacity~\cite{wang2024towards},
payload via meta-information~\cite{qu2025darkdistill}, transfer
frequency~\cite{kang2025efficient}, or mixed
proxy/knowledge/parameter sizes~\cite{le2025fedmekt}---preventing
aggregation.  \textbf{Qualitative claims are widespread}: 38
studies (75\,\%) assert communication savings, theoretically backed
by the $\mathcal{O}(MC)$ versus $\mathcal{O}(d)$ argument of
Section~\ref{sec:bg_kd}, but their empirical validation is uneven.
High-impact evidence outside the corpus corroborates the direction
of the claim: Wu et al.~\cite{wu2022communication} demonstrate, in a
widely cited \emph{Nature Communications} study, that a
distillation-based protocol exchanging a compact student model can
reduce federated communication cost by roughly an order of magnitude
without significant accuracy loss, a reference point against which
the byte-level reporting advocated here can be calibrated.

\subsubsection{Privacy assumptions, claims, and guarantees}
\label{sec:rq3_privacy}

Privacy claims span three levels of formality.  \textbf{Level 1:
structural privacy} ($\approx$30 studies)---raw data never leaves
the client, reinforced by the indirection of exchanging only soft
predictions or synthetic samples, as in
FedKD-CPI~\cite{wang2025fedkd}, meta-data
compression~\cite{yin2025meta}, and logit-limited graph
analysis~\cite{wang2025graph}.  These arguments are valid as far as
they go, but they are not formal guarantees.  Recent empirical
evidence sharpens the caveat considerably: Shi et
al.~\cite{shi2025unveiling} show that, in public-dataset-assisted
federated distillation, an honest-but-curious server can mount
label-distribution inference and likelihood-ratio membership
inference attacks using only the clients' predictions on the shared
proxy, achieving state-of-the-art attack performance against
\textsc{FedMD}-style frameworks; the soft labels exchanged on a
non-private proxy are therefore themselves a leakage channel.
\textbf{Level 2: explicit threat models without formal guarantees},
e.g., the two-tier information-loss argument of
FedSD2C~\cite{zhang2024one} and the high-level-feature claim
of~\cite{lu2024data}.  \textbf{Level 3: formal differential
privacy}---only 4~studies (8\,\%)~\cite{liang2023privacy,
liang2024enhanced, zheng2024fedgkd, jin2023feddyn}: LDP with
Gaussian noise before generator training (Precision 98.65
$\rightarrow$ 97.94 with DP)~\cite{liang2023privacy}, node-level DP
for graphs~\cite{zheng2024fedgkd}, and fake-ID strategies without a
bounded budget~\cite{jin2023feddyn}; stated budgets range over
$\epsilon \in \{0.5, 1, 2, 4, 8, 10\}$ with no consistent $\delta$
or composition reporting.  Privacy reporting correlates with proxy
type: generated-proxy studies provide substantive analysis more
often (8 of 22, 36\,\%) than public-reference studies (3 of 10),
because generative pipelines raise new memorization concerns that
authors are obliged to address.

\subsubsection{Summary of RQ3}

Four findings summarize RQ3.  First, the field under-reports its
non-functional dimensions (communication 31\,\%, privacy 33\,\%
substantive).  Second, performance gains over parameter averaging
are universal and grow with heterogeneity intensity, but protocol
heterogeneity prevents quantitative aggregation.  Third,
communication reporting is dominated by round counts, with only
18\,\% of studies providing byte-level metrics---the single largest
measurement gap.  Fourth, formal differential-privacy guarantees
appear in only 8\,\% of studies; the rest rely on structural
arguments that recent attacks show to be weaker than implied.

\subsection{RQ4: Proxy-quality criteria and reported metrics}
\label{sec:results_rq4}

\subsubsection{Qualitative criteria}
\label{sec:rq4_criteria}

Of the 51 studies, 49 (96\,\%) articulate at least one qualitative
criterion of proxy quality; two provide
none~\cite{lu2025domain, gao2022model}.  The leading criteria are
\textbf{diversity} (17~studies, 33\,\%), \textbf{alignment} with a
target distribution (11, 22\,\%), and \textbf{realism/fidelity} (8,
16\,\%), followed by coverage (6), class balance (5), and smaller
groups for confidence/reliability, formal privacy properties, size,
and domain-specific criteria---prototype and topology alignment in
federated graph learning~\cite{lin2025fedppd}, fault-mechanism
consistency~\cite{wang2023federated}, traffic-distribution
alignment~\cite{li2025flow}, and $V$-information
alignment~\cite{zhang2024one}.  The canonical triad of diversity,
alignment, and realism has thus informally consolidated, but it is
typically introduced as motivation rather than formalized into
measurable indicators, and the domain-specific refinements fragment
the vocabulary across areas.

\subsubsection{Quantitative metrics}
\label{sec:rq4_metrics}

Only 26~studies (51\,\%) report any numerical proxy-quality metric;
the other 25 fall back on downstream-task metrics that conflate
proxy quality with the rest of the pipeline.  The metrics that do
appear are inherited from centralized generative modeling---FID and
IS across diffusion stages~\cite{song2024fedadkd}, PSNR/SSIM with
$t$-SNE visualization~\cite{zhang2024one, zhang2023towards}---and do
not capture federation-specific concerns such as alignment with the
global label distribution.  Federation-specific metrics are rare and
idiosyncratic: precision-by-sample scores~\cite{lu2025fedfaa},
mutual-information leakage~\cite{yin2025meta}, information-gain and
data-integrity values~\cite{liu2025rapid}, and pseudo-graph--to-real
distance~\cite{lin2025fedppd}.  Most strikingly, despite diversity
being the most articulated criterion, almost no study reports an
explicit diversity metric---it is captured only implicitly through
IS or sample-count ablations.  Even without metrics, studies
operationalize quality through three heuristic families:
sampling/conditioning on an estimated target label
distribution~\cite{chen2025fedck, najafi2024enhancing,
liang2023privacy}; adversarial/regularization heuristics such as
Wasserstein training, diversity and transferability losses, and
EWC-style penalties against generator drift~\cite{liang2023privacy,
yi2025fedfld, zhang2023towards}; and filtering/selection (top-$K$,
core-set, semi-supervised, rejection sampling)~\cite{jin2023feddyn,
hao2024fedde, zhang2024one, karami2024gfd}---heuristics that are
rarely ablated to isolate their effect.

\subsubsection{Reported links between proxy quality and outcomes}
\label{sec:rq4_link}

Forty-three studies (84\,\%) claim that better proxy quality yields
better downstream performance, but only 14 (27\,\%) support the
claim with ablation or correlation evidence---e.g., positive
FID/IS-to-accuracy correlation across diffusion
stages~\cite{song2024fedadkd}, lower soft-target cross-entropy
linked to higher global accuracy~\cite{lu2025fedfaa}, and an
attribute-ablation showing that semantic-attribute-aware generation
is functionally superior on unseen fault
types~\cite{wang2023federated}.  Stability and convergence links are
discussed in 28~studies (55\,\%): warm-up phases stabilize
training~\cite{sun2022fed2kd}, regularization controls generator
shift~\cite{yi2025fedfld}, higher-quality proxies reduce
rounds-to-target~\cite{najafi2024enhancing, lu2024data}, and the
proxy acts as a convergence-smoothing regularizer under high
heterogeneity~\cite{le2023layer}.  The
mechanism$\to$quality$\to$outcome chain nonetheless remains
incomplete: no study answers ``if FID improves by $\Delta$, how much
does accuracy improve?''---leaving proxy-quality assessment an
artisanal practice rather than a measurable engineering discipline.

\subsubsection{Summary of RQ4}

Four findings summarize RQ4.  First, the literature has informally
converged on diversity, alignment, and realism as the dominant
criteria, with domain-specific refinements layered on top.  Second,
only 51\,\% of the corpus reports any quantitative proxy-quality
metric, and the metrics used are inherited from centralized
generative modeling.  Third, three heuristic
families---sampling/conditioning, adversarial/regularization, and
filtering/selection---translate criteria into algorithmic choices
but are rarely ablated.  Fourth, the quality-to-outcome chain is
supported by ablation evidence in only 27\,\% of studies.

\subsection{RQ5: Threat models, robustness, and defenses}
\label{sec:results_rq5}

\subsubsection{Threat landscape}
\label{sec:rq5_threats}

\begin{figure}[!htbp]
  \centering
  \includegraphics[width=.82\linewidth,height=.42\textheight,keepaspectratio]{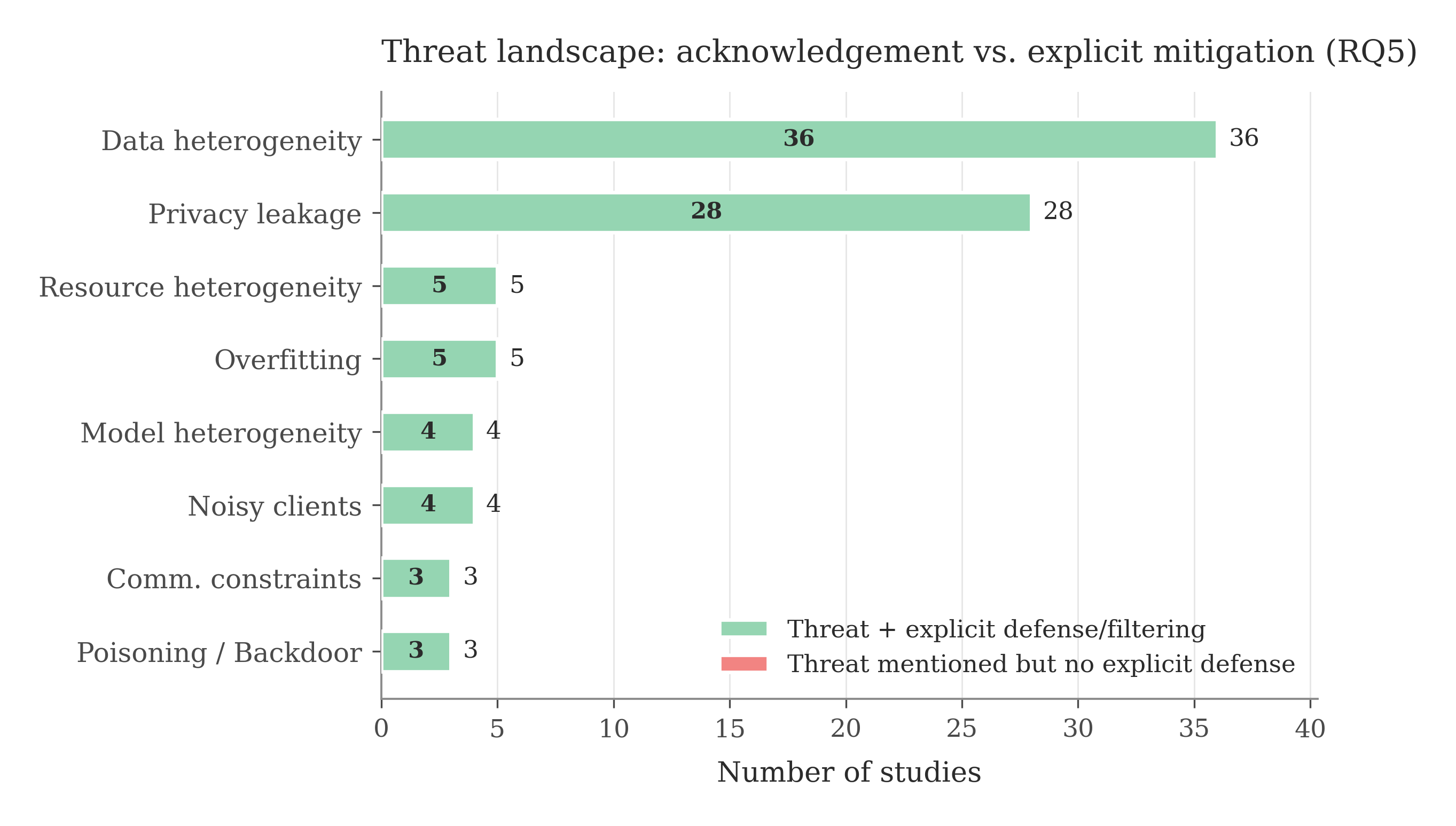}
  \Description{Matrix relating acknowledged threat categories to the defense mechanisms adopted.}
  \caption{Threats acknowledged across the corpus (RQ5) and the
    proportion of studies that pair the threat with an explicit
    defense or filtering mechanism.  Bar lengths show the number of
    studies that mention each threat; the green portion indicates
    studies that also report an explicit mitigation, while the
    coral portion indicates studies that mention the threat without
    a corresponding defense.}
  \label{fig:threat_defense_matrix}
\end{figure}

Figure~\ref{fig:threat_defense_matrix} shows the threats
acknowledged across the corpus and whether each is paired with an
explicit mitigation.  Robustness is the most uniformly addressed
non-functional dimension: 33~studies (65\,\%) provide substantive
discussion, 18 (35\,\%) treat it in passing, and none omits it.
The landscape is dominated by \textbf{data heterogeneity}
(36~studies, 71\,\%) and \textbf{privacy leakage} (28, 55\,\%),
followed by overfitting and resource heterogeneity (5 each,
10\,\%), noisy clients (4)~\cite{lu2025fedfaa, najafi2024enhancing,
song2024fedadkd, wang2022fedkg}, model heterogeneity as an
adversarial concern (4)~\cite{gao2022model, liang2023privacy,
sun2022fed2kd, zhang2023towards}, poisoning and backdoor attacks
(3)~\cite{liang2024enhanced, lu2025fedfaa, zhu2023adfl},
communication constraints (3)~\cite{kang2025efficient,
wang2022fedkg, zhang2023towards}, and two studies each for
catastrophic forgetting~\cite{yi2025fedfld, you2025adaptive}, model
drift~\cite{najafi2024enhancing, you2025adaptive}, and
out-of-distribution proxies~\cite{lu2024data, lu2025domain}.  The
recognized landscape is thus predominantly \emph{statistical}; the
adversarial threats central to the broader FL-security
literature---poisoning, backdoors, Byzantine clients---appear in
only 6\,\% of the corpus.  The poisoning end of the spectrum is
nonetheless well documented outside it: Fang et
al.~\cite{fang2020local} demonstrate local model-poisoning attacks
that defeat Byzantine-robust aggregation rules, a primary attack
reference against which the defenses below---including the
adversarial distillation defense of \textsc{ADFL}~\cite{zhu2023adfl}%
---can be positioned.

\subsubsection{Defenses and mitigation strategies}
\label{sec:rq5_defenses}

When a threat is acknowledged, a mitigation is almost always
proposed (49~studies, 96\,\%), falling into five families.
\textbf{Distillation-as-mitigation}: the claim that federated
distillation itself smooths heterogeneous teachers and avoids
sharing parameters---structurally valid but rarely supported by
attack-success comparisons.  \textbf{Adaptive aggregation and
weighting} (16~studies, 31\,\%): loss-inverse distillation
weights~\cite{song2024fedadkd}, relevance-score teacher
selection~\cite{lu2025fedfaa}, cosine-distance
weighting~\cite{zheng2024fedgkd}, inverse-entropy
averaging~\cite{karami2024gfd}, and mono-class logit
averaging~\cite{lin2025fedppd}---implicit Byzantine-robust
aggregators that downweight unreliable contributors.
\textbf{Differential privacy and noise injection}: the four formal
DP studies of Section~\ref{sec:rq3_privacy}, plus heuristic noise
against reconstruction~\cite{chen2025fedck} and DP combined with
permissioned channels~\cite{liang2024enhanced}.
\textbf{Synthetic-sample filtering}: relevance thresholds with
adaptive parameters~\cite{lu2025fedfaa}, top-$K$ selection with
distractors~\cite{jin2023feddyn}, and pseudo-label thresholds
(e.g., $\tau = 0.7$)~\cite{liang2023privacy}---effective when the
threat manifests as distributional artifacts.
\textbf{Architectural and protocol-level defenses}: global-model
purification in ADFL~\cite{zhu2023adfl}, QoS-based provider
filtering on permissioned blockchain~\cite{liang2024enhanced}, and
Fourier-perturbed distillates against membership inference and
inversion~\cite{zhang2024one}.

\subsubsection{Confidence filtering and temperature scaling}
\label{sec:rq5_filtering}

Two mechanisms recur across the corpus as robustness primitives even when
not framed as defenses.  \textbf{Temperature scaling}: 17~studies
(33\,\%) explicitly use it, 20 (39\,\%) explicitly do not (or use
$\tau = 1$), and 14 (27\,\%) do not specify; higher $\tau$ flattens
overconfident one-hot outputs on out-of-distribution inputs, yet
$\tau$ is almost always treated as a fixed hyperparameter inherited
from Hinton et al.~\cite{hinton2015distilling} rather than tuned for
robustness.  \textbf{Confidence-threshold filtering}: relevance
thresholds~\cite{lu2025fedfaa}, discrepancy
thresholds~\cite{yin2025meta}, top-$K$ rules~\cite{jin2023feddyn},
and pseudo-label cutoffs~\cite{liang2023privacy}, introduced against
unstable early-round generators but serving as a generic defense
against outlier contributors.  Used together---as in FedFAA,
FedAdKD, and Fed2KD---filtering removes degenerate samples and
temperature tempers the influence of the rest: the de-facto defense
against teacher unreliability.

\subsubsection{Gaps in adversarial-robustness evaluation}
\label{sec:rq5_gaps}

Three systemic gaps stand out.  \textbf{Out-of-distribution proxies
are rarely tested}: only 8~studies (16\,\%) handle them explicitly
---e.g., synthesis constrained to the intersection of local semantic
spaces~\cite{lu2024data} and noise-domain adaptation with
cross-domain discrepancy filtering~\cite{yin2025meta}---while the
remaining 84\,\% implicitly assume the proxy covers the union of
client distributions.  \textbf{Malicious-client handling is
essentially absent}: 3~studies (6\,\%), and partially so---QoS-based
filtering~\cite{liang2024enhanced} and indirect model
purification~\cite{zhu2023adfl}---in marked contrast with the
Byzantine-robust aggregation line of the broader FL literature.
\textbf{Quantitative attack success rates are almost never
reported}: no study implements a baseline attack, measures its
success rate without the defense, and re-evaluates with the defense
in place; claimed defenses therefore lack the empirical calibration
needed for cross-study comparison.

\subsubsection{Summary of RQ5}

Four findings summarize RQ5.  First, robustness is the most
uniformly discussed non-functional dimension (65\,\% substantive),
but the discussion concentrates on statistical rather than
adversarial threats.  Second, defenses are nearly universal (96\,\%)
yet dominated by the structural distillation-as-mitigation argument
plus adaptive aggregation and filtering.  Third, temperature scaling
and confidence filtering are the field's de-facto robustness
primitives.  Fourth, three systemic gaps persist: OOD proxies rarely
tested (16\,\%), malicious-client handling essentially absent
(6\,\%), and attack--defense evaluation missing across the corpus.

\subsection{RQ6: Methodological gaps and reporting practices}
\label{sec:results_rq6}

\subsubsection{Heterogeneity of non-IID experimental protocols}
\label{sec:rq6_protocols}

The operationalization of non-IID conditions is the single largest
barrier to cross-study comparison.  Dirichlet partitioning is the
de-facto standard ($\approx$38~studies) but with inconsistent
parameterization: $\alpha \in \{0.05, 0.1, 0.3, 0.5, 1.0, 10,
100\}$ across the corpus, with sweeps such as $\{0.01, 0.1,
1\}$~\cite{yi2025fedfld}, $\{0.05, 0.1, 1, 10\}$~\cite{sun2022fed2kd},
and $\{0.1, 0.3, 0.5\}$~\cite{zhang2024one}---so two studies
reporting ``Dirichlet non-IID'' may differ by orders of magnitude in
effective heterogeneity.  Non-Dirichlet protocols compound the
problem: two classes per device~\cite{le2023layer}, up-/down-sampling
of MNIST variants~\cite{wang2022fedkg}, class-coverage
ratios~\cite{zhang2023towards}, qualitative descriptions without
parameterization~\cite{wu2024efficient}, and naturally heterogeneous
institutional datasets in cross-silo work~\cite{lu2024hftl,
wang2025fedkd}.  Client counts additionally span two orders of
magnitude (2--179, median 10) and interact non-trivially with
$\alpha$; no study isolates the two effects.  Performance numbers
reported across studies---even on the same benchmark---are therefore
not directly comparable, making protocol standardization a
precondition for any future benchmark (Section~\ref{sec:future}).

\subsubsection{Benchmark fragmentation and external validity}
\label{sec:rq6_benchmarks}

The benchmark concentration of Section~\ref{sec:results_overview}
cuts both ways.  It creates methodological cohesion---accuracy on
CIFAR-10 is intuitively calibrated by all readers---but raises three
external-validity concerns documented across the corpus: methods
validated only on MNIST/CIFAR may not transfer to realistic image
scales (studies reporting Tiny-ImageNet or ImageNet observe
substantially smaller gains~\cite{zhang2024one, le2023layer,
najafi2024enhancing, wang2022fedkg}); methods validated only on
image classification may not transfer to other modalities, each
represented by only a handful of studies~\cite{gao2022model,
li2025flow, ding2024global, kang2025efficient, wang2025fedkd,
xu2024pefad, jin2023feddyn, bano2024fedcmd, le2025fedmekt,
lin2025fedppd, wang2025graph}; and the dominant benchmarks tolerate
weak teachers.  Twenty-eight studies (55\,\%) explicitly identify
the absence of non-image or harder-benchmark evaluation as the gap
they could not close.

\subsubsection{Quality-assessment profile of the corpus}
\label{sec:rq6_qa}

\begin{figure}[!htbp]
  \centering
  \includegraphics[width=.82\linewidth,height=.42\textheight,keepaspectratio]{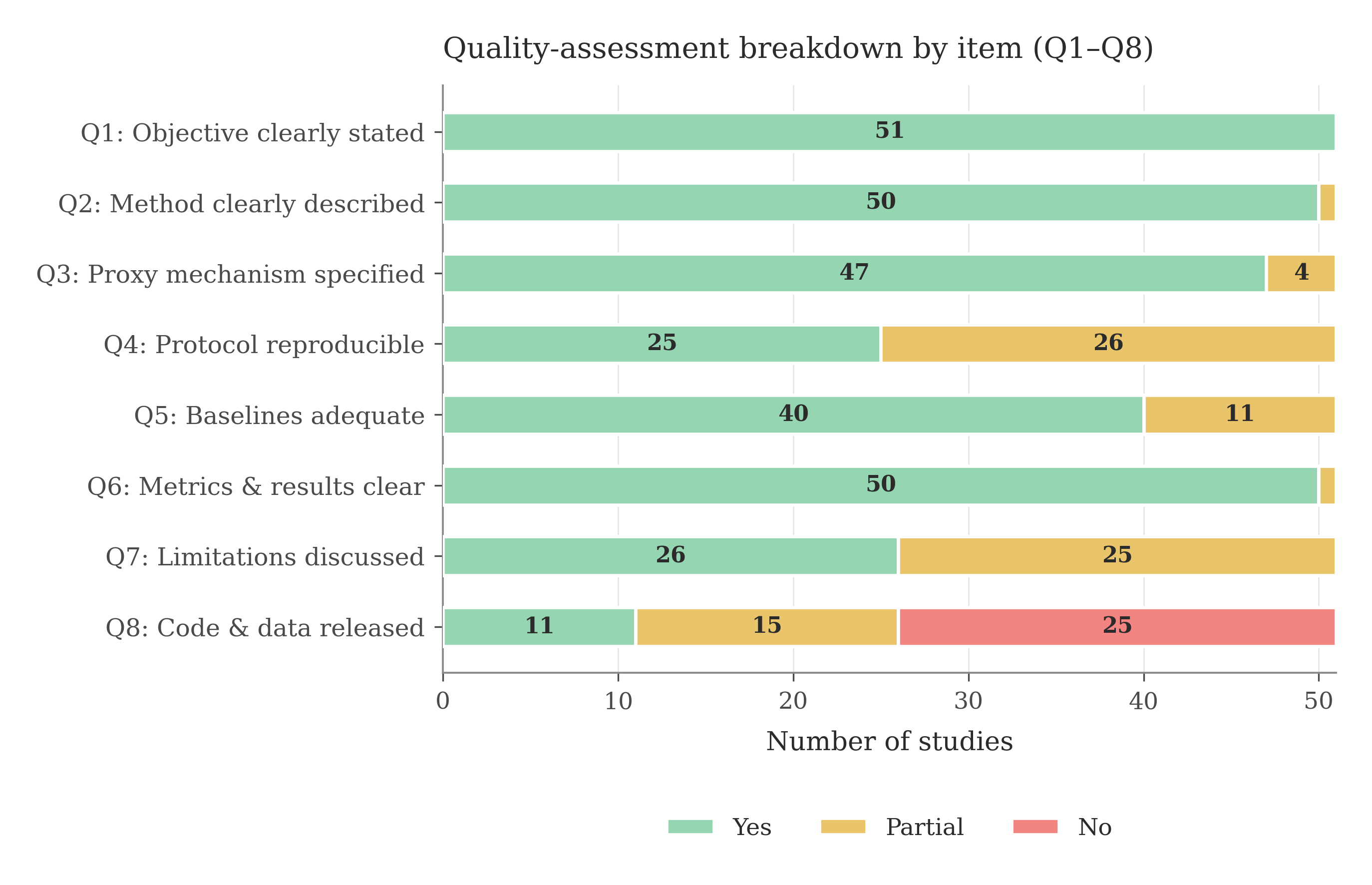}
  \Description{Stacked bar chart of Yes, Partial, and No answers per quality item.}
  \caption{Per-item quality-assessment breakdown: Yes (sage),
    Partial (gold), and No (coral) responses across the 51 included
    studies for each of the eight checklist items.  Q1, Q2, and Q6
    are nearly fully Yes; Q4 and Q7 split roughly evenly between
    Yes and Partial; Q8 is dominated by Partial and No responses.}
  \label{fig:qa_breakdown}
\end{figure}

Figure~\ref{fig:qa_breakdown}, together with Supplementary
Figs.~S7 and~S8, summarizes the quality profile scored by the
checklist of Section~\ref{sec:method_qa}.  Across the 51 studies,
the total score $S$ of Eq.~\eqref{eq:qa_total} has mean
$\bar{S} = 6.70$, median 7.00, standard deviation 0.83, and range
$[4.0, 8.0]$ (Q1 $=6.0$, Q3 $=7.0$); the distribution is unimodal
and right-shifted, with 26~studies (51\,\%) at or above 7.0 and only
5 (10\,\%) below 6.0.  Five items are essentially saturated: Q1
(objectives, mean 1.00), Q2 (method, 0.99), Q3 (proxy mechanism,
0.96), Q5 (baselines, 0.89), and Q6 (metrics, 0.99).  Three items
reveal the gaps: Q4 (protocol reproducibility, 0.75; 26 of 51
Partial), Q7 (limitations/robustness/privacy/communication
discussion, 0.75; 25 Partial), and Q8 (code and data release,
0.36---the weakest item by a large margin).  The field handles the
items internal to the paper well but under-handles those requiring
external validation---the classic profile of a young field whose
reproducibility engineering has not yet caught up.

\subsubsection{Reproducibility gap: code and synthetic-reference release}
\label{sec:rq6_reproducibility}

\begin{figure}[!htbp]
  \centering
  \includegraphics[width=.62\linewidth,height=.38\textheight,keepaspectratio]{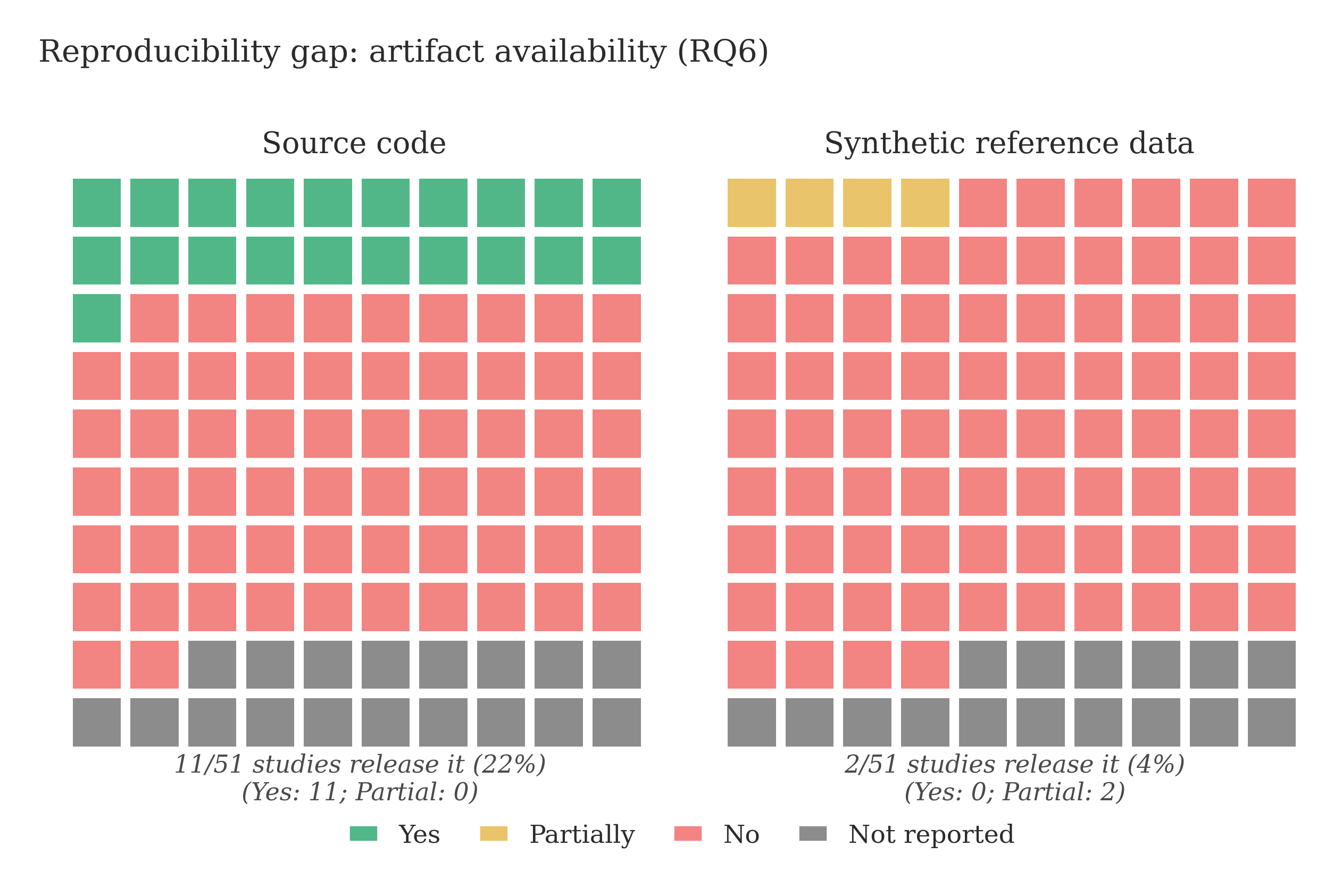}
  \Description{Waffle chart of artifact availability across the included studies.}
  \caption{Waffle chart of artifact availability across the
    corpus.  Each cell represents one percentage point of the
    51~included studies.  Source code is released by 11 studies
    (22\,\%); synthetic reference data is released, even partially,
    by only 2 studies (4\,\%).  No study in the corpus provides a
    complete release of both code and synthetic reference data.}
  \label{fig:reproducibility_waffle}
\end{figure}

Figure~\ref{fig:reproducibility_waffle} visualizes artifact
availability.  \textbf{Code}: 11~studies (22\,\%) release it, 31
(61\,\%) explicitly do not, and 9 (18\,\%) do not address the
question---unfavorable against the 50--70\,\% release rates
typically estimated for major ML venues, though the releasing
studies concentrate in recent years and reproducibility-minded
venues.  \textbf{Synthetic references}: \emph{no study releases its
synthetic reference data in full}; 2 (4\,\%) provide partial release
(generator checkpoints without materialized samples), 41 (80\,\%)
release nothing, and 8 (16\,\%) do not address it.  This is
particularly consequential because the proxy is the \emph{output} of
the process and is by construction non-private and shareable.
\textbf{The compounding effect}: only 2 studies provide both, even
partially, so a meticulous re-implementation would still be
evaluated on an uncontrolled re-generated proxy---the single largest
obstacle to cumulative empirical progress in this subfield.

\subsubsection{Consolidated methodological takeaway}
\label{sec:rq6_takeaway}

Combining Sections~\ref{sec:rq6_protocols}--\ref{sec:rq6_reproducibility}
with the cross-cutting observations from RQ1--RQ5, five persistent
gaps are the highest-leverage targets for near-term improvement.

\emph{Gap 1: standardized non-IID protocols.}  The proliferation of
Dirichlet $\alpha$ values, ad hoc partitions, and inconsistent
client counts prevents cross-study aggregation.  A community-agreed
canonical set---e.g., an $\alpha$ ladder $\{0.1, 0.5, 1.0\}$ on
client counts $\{10, 50, 100\}$---would enable comparable reporting
and meta-analysis-style synthesis.

\emph{Gap 2: bytes-to-convergence as a first-class metric.}
Communication efficiency motivates the field, yet only 18\,\% of
the corpus reports absolute byte-level metrics.  Bytes-to-%
convergence---the cumulative communication required to reach a
target accuracy---captures payload and convergence speed in a single
number and should become a mandatory reporting standard alongside
accuracy.

\emph{Gap 3: federation-aware proxy-quality metrics.}  Existing
metrics (FID, IS, PSNR, SSIM) are inherited from centralized
generative modeling and do not capture the alignment of the proxy
with the union of local distributions.  A federation-aware diversity
metric, calibrated against downstream performance, would close the
chain from \emph{mechanism} to \emph{quality indicator} to
\emph{outcome} (Section~\ref{sec:rq4_link}).

\emph{Gap 4: adversarial robustness with attack--defense
evaluation.}  The corpus contains no complete attack--defense
evaluation cycle (Section~\ref{sec:rq5_gaps}); defense claims remain
qualitative.  A community-curated catalogue of reference
attacks---membership inference, model inversion, backdoor injection
through the proxy, Byzantine teacher behavior---would enable
systematic robustness measurement.

\emph{Gap 5: transparent release of synthetic references.}  The
complete-release rate of synthetic proxy data is zero
(Section~\ref{sec:rq6_reproducibility}) despite being technically
non-controversial: the proxy is designed to be shareable.  Future
work should treat the synthetic reference as a first-class artifact,
on par with code, and venues should incorporate it into their
reproducibility checklists.

\subsubsection{Summary of RQ6}

Five findings summarize RQ6.  First, non-IID protocols are markedly
heterogeneous, with Dirichlet $\alpha$ spanning two orders of
magnitude.  Second, benchmark choice is fragmented, with a heavy
concentration on small image-classification datasets and a long
domain-specific tail.  Third, the quality profile is broadly
adequate (mean $\bar{S} = 6.70$, median 7.0) but splits sharply
between saturated internal items and weak external items.  Fourth,
the reproducibility gap is the corpus's weakest dimension: 22\,\%
code release and no complete synthetic-reference release.  Fifth,
five consolidated gaps---standardized protocols,
bytes-to-convergence, federation-aware metrics, attack--defense
evaluation, and transparent proxy release---define the agenda for
the field's next phase of maturation.

\section{Discussion}
\label{sec:discussion}

The synthesis of Section~\ref{sec:results} produced a dense factual
record of how 51 primary studies use synthetic and public proxies in
federated distillation.  This section consolidates the cross-cutting
implications along four axes: positioning, design implications,
heterogeneity and mismatch, and reproducibility.

\subsection{Positioning relative to existing surveys and frameworks}
\label{sec:disc_positioning}

Three points refine the positioning of Section~\ref{sec:related}
with substantive content.  First, \textbf{we confirm and quantify
what existing surveys assert qualitatively}: that public-reference
proxies are problematic and that the field is moving toward
generative alternatives becomes a measured trend---77\,\% of the
corpus uses generated, synthetic, or hybrid proxies, 75\,\%
evaluates under mixed heterogeneity, and 73\,\% appeared in
2024--2025.  Second, \textbf{we surface gaps that existing surveys
do not address}: the reproducibility diagnosis (22\,\% code release,
0\,\% complete synthetic-reference release) and the byte-based
communication analysis (18\,\%) are, to our knowledge, the first
quantitative measurements of these rates in the subfield, and they
justify the benchmark and artifact-release standards of
Section~\ref{sec:future}.  Third, \textbf{the synthetic-proxy axis
reorganizes the design space}: prior surveys organize methods by FL
challenge~\cite{wu2024fel_survey} or FL problem
type~\cite{qin2024fedkd_survey, ijcai2024fedkd_practical}, whereas making
the proxy mechanism the primary axis exposes the three structural
archetypes of Section~\ref{sec:rq1_patterns} that those axes hide.
The corpus's center of gravity has also shifted relative to earlier
surveys: GAN-based (25\,\%), hybrid (20\,\%), and other generative
(22\,\%) mechanisms are the modal practice today, while diffusion
and deep inversion together account for only 8\,\%.

\textbf{Two adjacent lines of work reinforce the trustworthiness
framing.}  \textsc{FedTKD}~\cite{chen2024fedtkd} is, to our
knowledge, the primary study that most explicitly shares this
review's framing: it combines adaptive knowledge distillation with
a threat-assessment mechanism that identifies malicious logit
contributions in heterogeneous FL, confirming that
\emph{trustworthy} federated distillation is consolidating as a
named research object.  Similarly,
\textsc{Selective-FD}~\cite{shao2024selective} addresses the regime
in which no well-trained global teacher is available---precisely the
open challenge documented in Section~\ref{sec:results_rq2}---through
selective knowledge sharing that filters misleading predictions on
both the client and the server side, with explicit
privacy-preserving intent.  Neither work was retrieved by our
protocol-driven query, which illustrates the recall limits inherent
to systematic term-based search; we therefore discuss them as
complementary context rather than as primary evidence.

\subsection{Implications for the design of federated distillation
  systems}
\label{sec:disc_design}

The structural patterns of Section~\ref{sec:results} translate into
five design recommendations.

\textbf{Recommendation 1: choose the proxy mechanism before the
distillation flow.}  Ownership scope is tightly coupled to proxy
type, and ownership in turn constrains the feasible flows
(Section~\ref{sec:rq1_patterns}); choosing the proxy first commits
the designer to a small set of coherent flow patterns, whereas
retrofitting a proxy onto a chosen flow rarely yields the same
coherence.

\textbf{Recommendation 2: align the proxy mechanism to the
deployment regime.}  Cross-device studies favor GAN-based and
hybrid mechanisms with server-side or hybrid ownership; cross-silo
studies dominate public references and feature-based distillation.
The reasons are structural---device clients cannot host generators,
and institutions can agree on shared references---so methods rarely
transfer across regimes without architectural change.

\textbf{Recommendation 3: budget for confidence filtering and
temperature scaling from day one.}  The field's de-facto robustness
primitives (Section~\ref{sec:rq5_filtering}) add hyperparameters
that interact with the whole pipeline; treating them as
afterthoughts produces methods that fail under noisy or
out-of-distribution proxies.  Both should be exposed as first-class
hyperparameters with reported tuning protocols.

\textbf{Recommendation 4: prefer hybrid proxies for cross-silo
deployments.}  Hybrid schemes systematically achieve the best
reported trade-offs in cross-silo settings---a public seed for
reliability, generative augmentation against residual
heterogeneity---while in cross-device deployments the per-client
overhead of the seed cancels the benefit.

\textbf{Recommendation 5: distinguish what distillation does from
what differential privacy does.}  The structural privacy argument
is valid but quantitatively weak: distillation reduces the leakage
attack surface but does not eliminate it, as numerous inference
attacks against soft labels have been demonstrated in the broader
privacy literature---most recently the label-distribution and
membership inference attacks of Shi et al.~\cite{shi2025unveiling}
against public-proxy federated distillation.  Practitioners
requiring formal guarantees should layer differential privacy on
top of the distillation pipeline; the four corpus studies that
combine the paradigms~\cite{jin2023feddyn, liang2023privacy,
liang2024enhanced, zheng2024fedgkd} provide templates.

\subsection{Impact of heterogeneity and proxy mismatch}
\label{sec:disc_mismatch}

The evidence on RQ3 (Section~\ref{sec:rq3_performance}) supports a
robust claim about proxy quality and performance under
heterogeneity, but with important qualifications.  Distillation is,
moreover, not the only family of remedies for non-IID degradation:
variance-reduction methods such as
\textsc{SCAFFOLD}~\cite{karimireddy2020scaffold} correct client
drift through control variates, and contrastive approaches such as
\textsc{MOON}~\cite{li2021moon} regularize local training against
the global representation---both without any proxy data.  The
comparative advantage of proxy-based distillation must therefore be
argued against these parameter-space alternatives, chiefly on the
grounds of model heterogeneity and communication profile rather
than non-IID robustness alone.

\textbf{Gains scale with heterogeneity intensity, but not for
free}: the advantage over \textsc{FedAvg} grows from
$\approx$1--3\,\% at $\alpha = 1.0$ to $\approx$8--30\,\% at
$\alpha \le 0.1$, yet the generator must itself be trained and
tuned, and hyperparameter sensitivity grows with heterogeneity---at
$\alpha = 0.05$ the same generator may converge or diverge depending
on seeds, balance regularization, and activation
timing~\cite{sun2022fed2kd, yi2025fedfld}.  \textbf{Proxy mismatch
is the dominant under-tested failure mode}: only 16\,\% of studies
explicitly handle out-of-distribution proxies
(Section~\ref{sec:rq5_gaps}), although it is precisely the mismatch
between proxy and global mixture that makes distillation work or
fail under strong heterogeneity.  \textbf{The cross-silo case
deserves special attention}: feature skew is structural in
institutional data, formal privacy matters most there, yet only
3~studies evaluate feature skew---an under-served niche at the
intersection of strong feature skew, formal privacy, and small
client populations.  Finally, \textbf{mechanism choice is not
orthogonal to heterogeneity intensity}: public references degrade
gracefully but collapse at extreme skew; GANs can compensate but
grow unstable; diffusion's fidelity advantage is offset by
per-round retraining cost.  No mechanism is uniformly best.

\subsection{Reproducibility, transparency, and availability of
  proxies}
\label{sec:disc_repro}

The findings of Section~\ref{sec:rq6_reproducibility} qualify the
interpretation of every other finding.  When 80\,\% of studies do
not release their synthetic reference, reported numbers cannot be
verified or extended: even a meticulous re-implementation is
evaluated on an uncontrolled re-generated proxy---a qualitatively
different reproducibility problem from centralized ML, where
releasing model and dataset suffices.  The 22-percentage-point
asymmetry between code release (22\,\%) and complete proxy release
(0\,\%) indicates that authors do not yet conceptualize the
synthetic reference as a publishable artifact; the contributing
factors---storage cost and the absence of an agreed sharing
format---are both tractable.  Synthetic-reference release is
technically non-controversial: unlike model weights, the proxy is by
construction non-private and intended for distribution, so the
current 0\,\% rate is a cultural artifact, not a technological
constraint.  We argue in Section~\ref{sec:future} that the synthetic
reference should become a first-class artifact with a versioned
identifier, generator metadata, and a re-use license.  Consistent
with this position, the complete extraction workbook of this review
is itself publicly archived on Zenodo under a versioned
DOI~\cite{freire2026workbook}.

\subsection{Limitations of the evidence base}
\label{sec:disc_limitations}

The conclusions of this review are constrained not only by its own
methodological choices (Section~\ref{sec:method_threats}) but by the
design and reporting practices of the primary studies.
\textbf{Modest experimental scale}: median 10 clients, only 5
studies beyond 50; the $10^3$--$10^6$ cross-device populations that
motivate the field~\cite{kairouz2021advances} are unrepresented, so
gains observed at small scale may not extrapolate, and rankings at
$K = 10$ may differ at $K = 1000$.  \textbf{Non-standardized
non-IID protocols}: even at the same nominal Dirichlet parameter,
differences in partition counts, seed handling, and schedules can
shift effective heterogeneity by more than the method-to-method
differences reported; our synthesis therefore relies on patterns of
direction rather than pooled effect sizes.  \textbf{Heterogeneous
and frequently weak baselines}: comparisons range from
\textsc{FedAvg} alone to state-of-the-art federated-distillation
suites~\cite{mcmahan2017fedavg, li2020fedprox,
jeong2018communication, lin2020ensemble, zhu2021datafree},
so method-to-method advantages measure different quantities across
studies (Q5 mean $\bar{s} = 0.89$).  \textbf{Incomplete reporting of
secondary dimensions}: communication substantively analyzed in
31\,\% of studies, formal privacy in 8\,\%, quantitative proxy
quality in 51\,\%, adversarial evaluation essentially absent.  The
combined effect is that this review can characterize the
\emph{shape} of the field with confidence, but cannot deliver
finely calibrated quantitative recommendations about which method
outperforms which under which precise conditions---the
methodological investments of Section~\ref{sec:future} are what
would make such recommendations possible.

\section{Future Research Directions}
\label{sec:future}

The synthesis converges on a research agenda along seven directions:
five methodological investments that close the gaps consolidated in
Section~\ref{sec:rq6_takeaway}, and two forward-looking extensions.
Each direction is presented as gap, agenda, and progress criteria.

\subsection{Standardized non-IID benchmarks for federated
  distillation}
\label{sec:future_benchmarks}

\emph{Gap} (Section~\ref{sec:rq6_protocols}): Dirichlet $\alpha$
values span two orders of magnitude, ad hoc partitions coexist with
parametric ones, and client counts range from 2 to 179 with no
standard configuration.  \emph{Agenda}: a community-curated
benchmark suite organized around (i)~a canonical $\alpha$ ladder,
e.g.\ $\alpha \in \{0.1, 0.5, 1.0\}$ for severe, moderate, and mild
skew, reported in full by any new method; (ii)~a canonical
client-count ladder, e.g.\ $K \in \{10, 50, 100\}$ with optional
$K = 1000$ stress testing, with cross-device claims requiring at
least the medium and large scales; and (iii)~a multi-modality core
that adds at least one tabular, one time-series, and one graph or
NLP task to the canonical image benchmarks.  \emph{Progress}: a
published benchmark specification with reference implementations,
adoption tracking in new papers, and the convergence of cross-study
numbers that would enable retrospective meta-analysis.

\subsection{Bytes-to-convergence as a first-class metric}
\label{sec:future_btc}

\emph{Gap} (Section~\ref{sec:rq3_comm}): 75\,\% of the corpus claims
communication savings, but only 18\,\% reports bytes; rounds hide
the per-round payload.  \emph{Agenda}: adopt
\emph{bytes-to-convergence} (BTC) as a mandatory reporting standard,
defined as the cumulative bytes exchanged to reach a target metric
value $\eta$:

\begin{equation}
  \label{eq:btc}
  \mathrm{BTC}(\eta)
  \;=\;
  \sum_{t=1}^{T(\eta)}
    \sum_{k \in \mathcal{S}_t}
    \mathrm{payload}_{t,k},
\end{equation}

where $T(\eta)$ is the first round at which the global model reaches
$\eta$ and $\mathrm{payload}_{t,k}$ is the byte count of the
client--server exchange in round $t$.  BTC composes payload and
convergence speed in a single number comparable across methods
regardless of what they exchange; fixing $\eta_{\mathrm{target}}$
per canonical benchmark (e.g., 80\,\% on CIFAR-10) yields a simple
standard variant, with per-class or bandwidth-capped refinements as
needed.  \emph{Progress}: BTC in venue checklists, reporting rates
in new papers, and the emergence of cross-study
(BTC,~accuracy) Pareto frontiers.

\subsection{Federation-aware proxy-quality metrics}
\label{sec:future_metrics}

\emph{Gap} (Sections~\ref{sec:rq4_metrics}--\ref{sec:rq4_link}):
the metrics in use (FID, IS, PSNR, SSIM) are inherited from
centralized generative modeling and miss federation-specific
properties, and the quality-to-outcome chain is supported by
ablations in only 27\,\% of studies.  \emph{Agenda}: three concrete
directions.  First, a \emph{federation-aware diversity metric}
measuring how proxy coverage distributes over the client-conditional
marginals, e.g., the average maximum mean discrepancy between the
proxy and each client's empirical distribution:

\begin{equation}
  \label{eq:fed_diversity}
  D_{\mathrm{fed}}(\mathcal{D}_{\mathrm{prx}})
  \;=\;
  \frac{1}{K} \sum_{k=1}^{K}
    \mathrm{MMD}^2(\mathcal{D}_{\mathrm{prx}}, \mathcal{D}_k),
\end{equation}

estimated with kernel statistics on aggregate features clients can
disclose; lower $D_{\mathrm{fed}}$ indicates better coverage of the
union of local distributions.  Second, a \emph{calibration of
quality to outcome}: systematically vary one quality dimension and
measure the accuracy response, producing curves that translate
``$\Delta$\,FID'' into ``$\Delta$\,accuracy'' and enable
proxy-quality engineering rather than guesswork.  Third, a
\emph{coverage diagnostic} that flags per-class or per-cluster gaps
before training, preventing the silent failures of
Section~\ref{sec:disc_mismatch}.  \emph{Progress}: a validated
federation-aware metric, calibration curves in published methods,
and diagnostics in reference implementations.

\subsection{Adversarial robustness with attack--defense evaluation}
\label{sec:future_robustness}

\emph{Gap} (Section~\ref{sec:rq5_gaps}): the corpus contains no
complete attack--defense evaluation cycle, so defense claims remain
qualitative and incomparable.  \emph{Agenda}: a community-curated
catalogue of reference attacks with companion evaluation protocols,
covering four categories---(1)~membership inference against the
soft labels; (2)~model inversion against the proxy;
(3)~backdoor injection through the proxy pipeline or a server-side
generator; and (4)~Byzantine teacher behavior.  For each attack the
catalogue should specify the threat model, a reference
implementation, baseline success rates on a vanilla
federated-distillation pipeline, and the expected effect of the
common defenses (differential privacy at stated $\epsilon$,
confidence filtering, robust aggregation).  Recent work provides
natural starting points: the proxy-based membership and
label-distribution inference attacks of Shi et
al.~\cite{shi2025unveiling} for the first category, the
logit-credibility assessment of
\textsc{FedTKD}~\cite{chen2024fedtkd} and the selective knowledge
sharing of \textsc{Selective-FD}~\cite{shao2024selective} for the
defense side, and the model-poisoning attacks of Fang et
al.~\cite{fang2020local} for the Byzantine category.
\emph{Progress}: catalogue publication, attack-success-rate
reporting in new methods, and cross-study defense comparisons
replacing unsubstantiated claims.

\subsection{Transparent release of synthetic proxies}
\label{sec:future_release}

\emph{Gap} (Section~\ref{sec:rq6_reproducibility}): no corpus study
completely releases its synthetic reference, a cultural rather than
technical constraint.  \emph{Agenda}: treat the synthetic reference
as a first-class artifact released alongside the paper, in a
community-agreed format with four elements: (a)~the materialized
proxy samples in a standard serialization; (b)~metadata describing
the generator (architecture, hyperparameters, seeds, training-data
summary); (c)~the global model's soft-label predictions on the proxy
at the end of training, as the oracle against which
re-implementations validate; and (d)~a permissive license enabling
re-use.  Storage requirements are well within standard repositories
(Zenodo, Figshare, HuggingFace Datasets), and versioned identifiers
would allow unambiguous reference and re-use tracking.
\emph{Progress}: release rates in new papers, integration into
venue reproducibility checklists, and the emergence of cross-paper
proxy comparisons enabling cumulative empirical progress.

\subsection{Architectural co-design of generator and distillation}
\label{sec:future_codesign}

The literature treats the generator and the distillation step as
independent components: the generator is judged by generic criteria
(FID, IS) that do not align with what distillation actually
needs---which depends on the global model's current decision
boundary and the clients' heterogeneity profile.  Co-designed
architectures would couple the two in three ways:
\emph{distillation-aware generator training}, optimizing the
generator jointly with a proxy of the distillation loss so it
produces samples that maximize the information content of the
resulting soft labels; \emph{round-aware generation}, scheduling
harder samples in late rounds and easier ones early, tracking the
evolving decision boundary; and \emph{client-aware generation},
producing customized proxy subsets per client or group, tailored to
gaps in their local distributions and connected to the client-side
ownership scope of Section~\ref{sec:rq1_ownership} and the
personalized-FL literature.  Progress would be measured by methods
that jointly optimize a co-design objective and beat the decoupled
baseline on standardized benchmarks.

\subsection{Extension to non-image modalities}
\label{sec:future_modalities}

The corpus concentrates on image classification (47\,\%), and
image-tuned methods do not transfer directly: diffusion suits
tabular and time-series synthesis poorly, and GANs struggle with
sparse or discrete inputs.  Three directions have high value.
First, \emph{tabular and time-series synthesis with VAE/AE
backbones}: the three corpus studies targeting non-image data
(PeFAD~\cite{xu2024pefad}, Fed2KD~\cite{sun2022fed2kd},
FedSD2C~\cite{zhang2024one}) all rely on VAE or autoencoder
mechanisms, and systematic study on tabular and time-series
benchmarks would fill a clear gap.  The broader FL literature
confirms that the time-series setting is active and practically
relevant---personalized federated frameworks for passenger-flow
prediction~\cite{torrepadula2025fedflow} and interpretable federated
multivariate time-series classification~\cite{younis2023flames2graph}
are recent examples---but neither line yet incorporates distillation
with synthetic proxies, underscoring the gap identified here.
Second, \emph{graph-structured synthesis}: FedPPD~\cite{lin2025fedppd}
and FedGKD~\cite{zheng2024fedgkd} are isolated examples of a line
that raises structural questions (which sub-graphs to share, which
properties to preserve, node- versus graph-level privacy) deserving
systematic treatment.  Third, \emph{multimodal proxies}: cross-modal
distillation, illustrated by FedCMD~\cite{bano2024fedcmd} and
FedMEKT~\cite{le2025fedmekt}, poses the underexplored problem of
synthesizing aligned multimodal samples, with high relevance for
embodied-AI and edge applications.  Progress would be measured by
modality-specific method lines with their own benchmarks, and by
cross-modality comparisons testing whether a single framework
generalizes or each modality requires its own treatment.

\subsection{Summary of the research agenda}
\label{sec:future_summary}

The first five directions are methodological investments that close
measurable gaps---standardized benchmarks, bytes-to-convergence,
federation-aware metrics, attack--defense evaluation, and
transparent proxy release---and would make cumulative empirical
progress possible; the last two are extensions that emerge from the
corpus's empirical patterns.  Their convergence over the next three
to five years will be the principal determinant of whether federated
distillation with synthetic proxies becomes a deployed technology or
remains a productive but isolated research area.

\section{Conclusion}
\label{sec:conclusion}

This systematic literature review examined how synthetic proxies
are employed for knowledge distillation in federated learning,
drawing on 51~primary studies (2020--2025) retrieved from five
major databases under the protocol of
Section~\ref{sec:methodology}.  Five findings characterize the
field.  It is young and fast-growing, with 73\,\% of the corpus
published in 2024--2025.  It has moved decisively away from the
assumption of a pre-existing public reference, with 77\,\% of
studies using generated, synthetic, or hybrid proxies.  Its design
space organizes into a small number of recurring archetypes that
couple proxy mechanism, distillation flow, and ownership scope.
Its experimental practice remains dominated by small-scale
image-classification benchmarks.  And, most consequentially, its
reproducibility posture is weak: only 22\,\% of studies release
source code, and no study completely releases the synthetic
reference data---an artifact that, by construction, is designed to
be shareable.

\textbf{Contributions.}  The review contributes four artifacts to
the community.  (i)~A \textbf{three-axis taxonomy} of synthetic
proxies in federated distillation (proxy mechanism, distillation
flow, ownership scope) that exposes structural patterns hidden by
prior FL-challenge or FL-problem-based classifications.  (ii)~A
\textbf{quantitative evidence base} for cross-cutting claims
about the field, replacing narrative assertions with measured
percentages on a transparent corpus.  (iii)~A \textbf{diagnostic
of the reproducibility gap} that, to our knowledge, is the first
quantitative measurement of code- and synthetic-reference release
rates in this subfield, identifying a 22-percentage-point gap
between the two and providing a measurable target for community
improvement.  (iv)~A \textbf{seven-direction research agenda}
(Section~\ref{sec:future}) that consolidates the methodological
gaps documented in Section~\ref{sec:results_rq6} into actionable
investments---standardized non-IID benchmarks, bytes-to-convergence
reporting, federation-aware quality metrics, attack--defense
evaluation, and transparent proxy release---together with two
forward-looking opportunities (architectural co-design and
extension to non-image modalities).

\textbf{Implications for research.}  The review's central message
for the research community is that the field is now mature enough
to benefit from infrastructure investments rather than from
additional methodological inventiveness alone.  The proliferation
of methods that work qualitatively---and the absence of
quantitative comparability among them---means that the marginal
return on the next new method is low compared to the marginal
return on a community-curated benchmark, a community-agreed
artifact-release format, or a reference catalogue of attacks.  We
therefore encourage venues with strong reproducibility cultures to
incorporate synthetic-reference release into their
artifact-availability checklists, and we encourage authors to
treat the synthetic proxy as a first-class deliverable on par
with the source code.

\textbf{Implications for practice.}  For practitioners selecting
or designing methods for deployment, the review's structural
patterns translate into a small set of decision rules: choose the
proxy mechanism before the distillation flow; align the mechanism
to the deployment regime (cross-device versus cross-silo); budget
for confidence filtering and temperature scaling from the start;
prefer hybrid proxies for cross-silo deployments; and layer
formal differential privacy on top of distillation rather than
relying on distillation's structural privacy alone.  These rules,
articulated in Section~\ref{sec:disc_design}, narrow the
practitioner's choice from a combinatorially large design space
to a small number of coherent archetypes.

The complete extraction workbook---including bibliographic
metadata, harmonized categorical fields, raw textual extractions,
and per-item quality scores for every included study---is publicly
archived with a persistent DOI~\cite{freire2026workbook} and is the
primary resource for any extension or replication of this review.
Looking forward, we believe that the convergence of standardized
benchmarks, transparent artifact release, and federation-aware
quality metrics over the next three to five years will be the
principal determinant of whether synthetic and public proxies for
federated distillation transition from a productive research area
into a \emph{trustworthy} deployed technology.

\textbf{Limitations.}  The review's conclusions remain constrained
by the design and reporting practices of the primary studies
themselves (Section~\ref{sec:disc_limitations})---modest
experimental scale, non-standardized non-IID protocols,
heterogeneous baselines, and incomplete reporting of secondary
dimensions---as well as by the deliberate restriction of the corpus
to peer-reviewed venues, which together bound the strength of the
cross-study claims made here.


\begin{acks}
This study was financed in part by the Coordena\c{c}\~{a}o de
Aperfei\c{c}oamento de Pessoal de N\'{\i}vel Superior -- Brasil
(CAPES) -- Finance Code 001, and the Brazilian agencies FACEPE and
CNPq.
\end{acks}

\bibliographystyle{ACM-Reference-Format}
\bibliography{sn-bibliography}

\end{document}